\documentclass[letterpaper, 10 pt, conference]{ieeeconf}  

\IEEEoverridecommandlockouts                              

\usepackage{algorithm}
\usepackage{algpseudocode}

\usepackage{graphics} 
\usepackage{epsfig} 
\usepackage{epstopdf}
\usepackage{mathptmx} 
\usepackage{times} 
\usepackage{amsmath} 
\usepackage{amssymb}  
\usepackage{bm}
\usepackage{upgreek}
\usepackage{subfigure}
\usepackage{multirow}
\usepackage{float}
\usepackage{url}
\usepackage{textcomp}
\usepackage{color}

\newtheorem{theorem}{Theorem}
\newtheorem{lemma}{Lemma}
\newtheorem{procedure}{Procedure}
\newtheorem{definition}{Definition}
\newtheorem{note}{Note}

\newtheorem{approach}{Approach}

\title{Human-Machine Co-Adaptation for Robot-Assisted Rehabilitation via Dual-Agent Multiple Model Reinforcement Learning (DAMMRL)}

\author{Yang An$^1$ $^2$, Yaqi Li$^1$, Hongwei Wang$^1$, Rob Duffield$^3$, Steven W. Su$^1$ $^2$*	
\thanks{$^{*}$ The co-responding author.}
\thanks{$^{1}$ Jinan Key Lab of Intelligent Rehabilitation Robotics, College of Artificial Intelligence and Big data for Medical Sciences, Shandong First Medical University
Shandong Academy of Medical Sciences.}
\thanks{$^{2}$ Faculty of Engineering and IT, University of Technology Sydney, NSW, 2007, Australia}
\thanks{$^{3}$ Faculty of Health, University of Technology Sydney, NSW, 2007, Australia}
}

\begin{document}

\maketitle
\thispagestyle{empty}
\pagestyle{empty}

\begin{abstract}
	This study introduces a novel approach to robot-assisted ankle rehabilitation by proposing a Dual-Agent Multiple Model Reinforcement Learning (DAMMRL) framework, leveraging multiple model adaptive control (MMAC) and co-adaptive control strategies. In robot-assisted rehabilitation, one of the key challenges is modelling human behaviour due to the complexity of human cognition and physiological systems. Traditional single-model approaches often fail to capture the dynamics of human-machine interactions. Our research employs a multiple model strategy, using simple sub-models to approximate complex human responses during rehabilitation tasks, tailored to varying levels of patient incapacity. The proposed system's versatility is demonstrated in real experiments and simulated environments. Feasibility and potential were evaluated with 13 healthy young subjects, yielding promising results that affirm the anticipated benefits of the approach. This study not only introduces a new paradigm for robot-assisted ankle rehabilitation but also opens the way for future research in adaptive, patient-centred therapeutic interventions.
\end{abstract}

\section{INTRODUCTION}
Advancements in robotics have significantly enhanced robot-assisted rehabilitation, helping patients regain motor functions through innovative therapies \cite{khalid2023robotic}. Robotic devices provide precise movement guidance and adjustable resistance, supporting exercises for muscle strengthening and joint flexibility \cite{naqvi2024dual}. Recent developments focus on improving systems for both upper and lower limb rehabilitation, using technologies like virtual reality and game-like scenarios to make therapy more engaging and effective. These innovations are crucial for better patient outcomes, mimicking natural movements and providing essential feedback to improve motor skills.



Fundamentally, when considering the rehabilitation purpose for different stages of the rehabilitation (e.g., active rehabilitation and passive rehabilitation),  the robot-assisted rehabilitation process can be regarded as a complicated sequential decision-making process \cite{stevanovic2022joint} \cite{zhang2022reinforcement} \cite{mukherjee2022survey}, for which the Reinforcement Learning (RL) based approaches are often employed as a powerful tool to perform its optimization process. 
Incorporating reinforcement learning (RL) into rehabilitation also naturally leverages the human feedback loop. The reward prediction error (RPE) theory of dopamine function, rooted in RL, differentiates goal-directed from habitual learning strategies \cite{doll2012ubiquity}. Given the subconscious and individual nature of human learning, personalized evaluation is crucial for optimizing human-machine interfaces (HMI) using RL \cite{doll2012ubiquity} \cite{zini2022adaptive} \cite{ghadirzadeh2016sensorimotor}. Furthermore, Model-Based RL (MBRL) is preferred for human decision-making, aligning with dual process theory and the 'thinking fast and slow' framework \cite{plaat2020model} \cite{HAMRICK20198} \cite{kahneman2011before}.

In robot-assisted rehabilitation, to effectively supervise the adaptation process,  in clinical practice, human-machine co-adaptation is essential through developing mutual learning algorithms \cite{kubota2022methods}. To this end, this paper introduces a patient-machine dual-agent decision process that emphasizes human adaptation to the machine, building on our recent research \cite{guo2024cooperative}. In that work, we proposed a Cooperative Adaptive Markov Decision Process (CAMDP) model to study patient-machine co-adaptation, focusing on the convergence and switching behavior of both agents as an initial step in developing policy adaptation algorithms \cite{bai2019provably}. Additionally, we modified the finite MDP-based reinforcement learning approach to address non-stationarity in multi-agent reinforcement learning (MARL).

The main purpose of introducing the dual-agent MDP model is to enhance the performance and efficiency of human-machine systems through RL-based learning in both online and offline scenarios \cite{guo2024cooperative}. Learning efficiency is measured by the human's proficiency in controlling the machine. In machine-oriented reinforcement learning, research often includes learning curves that plot optimality (cumulative return) on the y-axis against various measures of time complexity (such as real-world sample complexity, model sample complexity, and computational complexity) on the x-axis. Studies suggest that RL-based reinforcement learning can effectively enhance both cumulative reward and optimality. In this study, we designed reward functions that incorporate desired exercise performances and rehabilitation effect indicators. These functions are tailored to different rehabilitation purposes and stages of the patient, allowing for patient-oriented personalized robot-assisted rehabilitation. 

The scarcity of clinical data and the need for safety and explainability \cite{moerland2020model} make Model-based MARL \cite{yang2020overview} ideal for clinical studies like robot-assisted rehabilitation. It demonstrates advantages in data utilization, human knowledge integration, safety, and explainability \cite{moerland2020model}. State and temporal abstraction can further leverage these benefits.
This is why, in \cite{guo2024cooperative}, we investigated two agents with finite states and actions described by finite MDPs \cite{qingji2008robot} \cite{wang2007emotion}.

However, there are many difficulties in using a fully finite MDP model-based multi-agent RL with finite actions to address real application issues, especially in robot-assisted rehabilitation, where both the state and action spaces are often infinite. Additionally, real-time model identification adds computational burdens. In this study, to apply the approach given in \cite{guo2024cooperative} with finite states and actions described by finite MDPs \cite{qingji2008robot} \cite{wang2007emotion}, we proposed multiple model approaches used on our previous research \cite{8880648} \cite{huang2021human}. 

\begin{figure}[ht]
	\centering
	\vspace*{-0.1cm} 
	\hspace*{0.1cm}\includegraphics[width=0.45\textwidth]{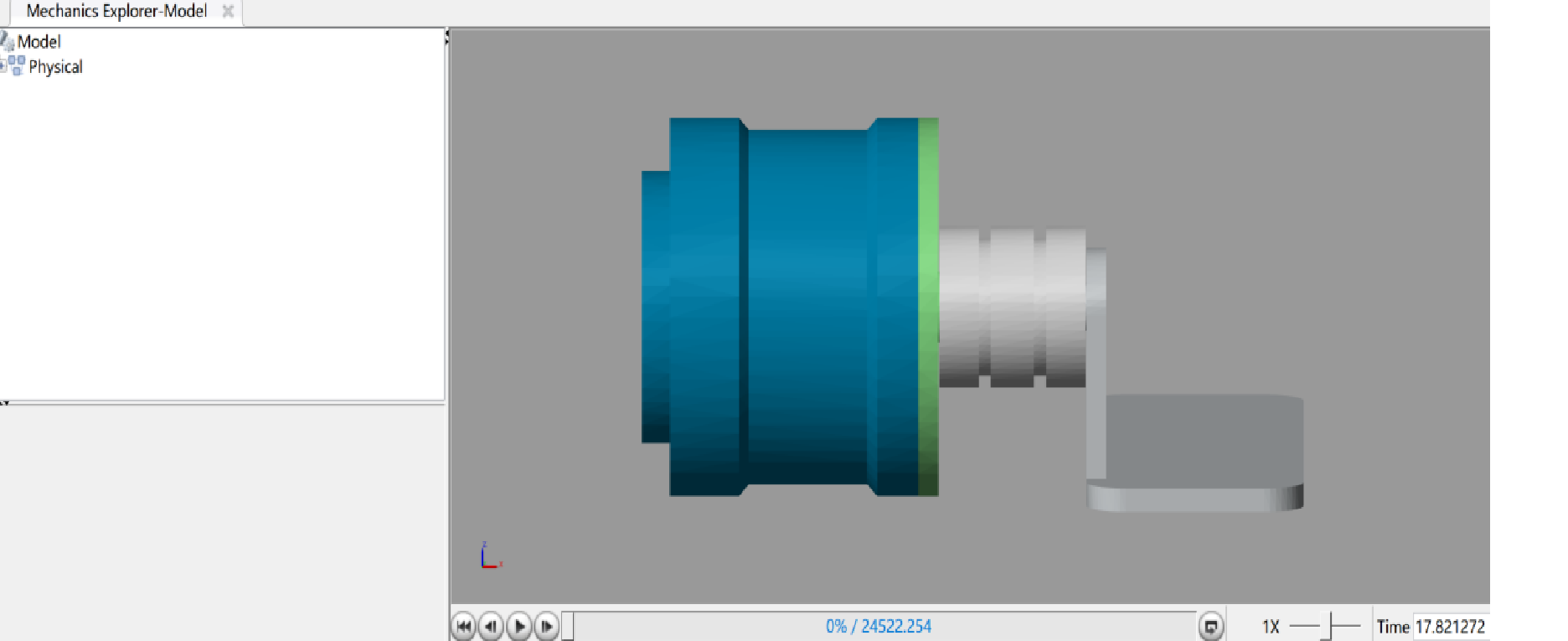}
	\vspace*{-0.cm} 
	\caption{The robot's model in Simulink/SimMechanics used for RL training.}
	\label{robot_simulink}
\end{figure}

\begin{figure}[ht]
	\centering
	\vspace*{0.0cm} 
	\hspace*{-0.1cm}\includegraphics[width=0.65\textwidth]{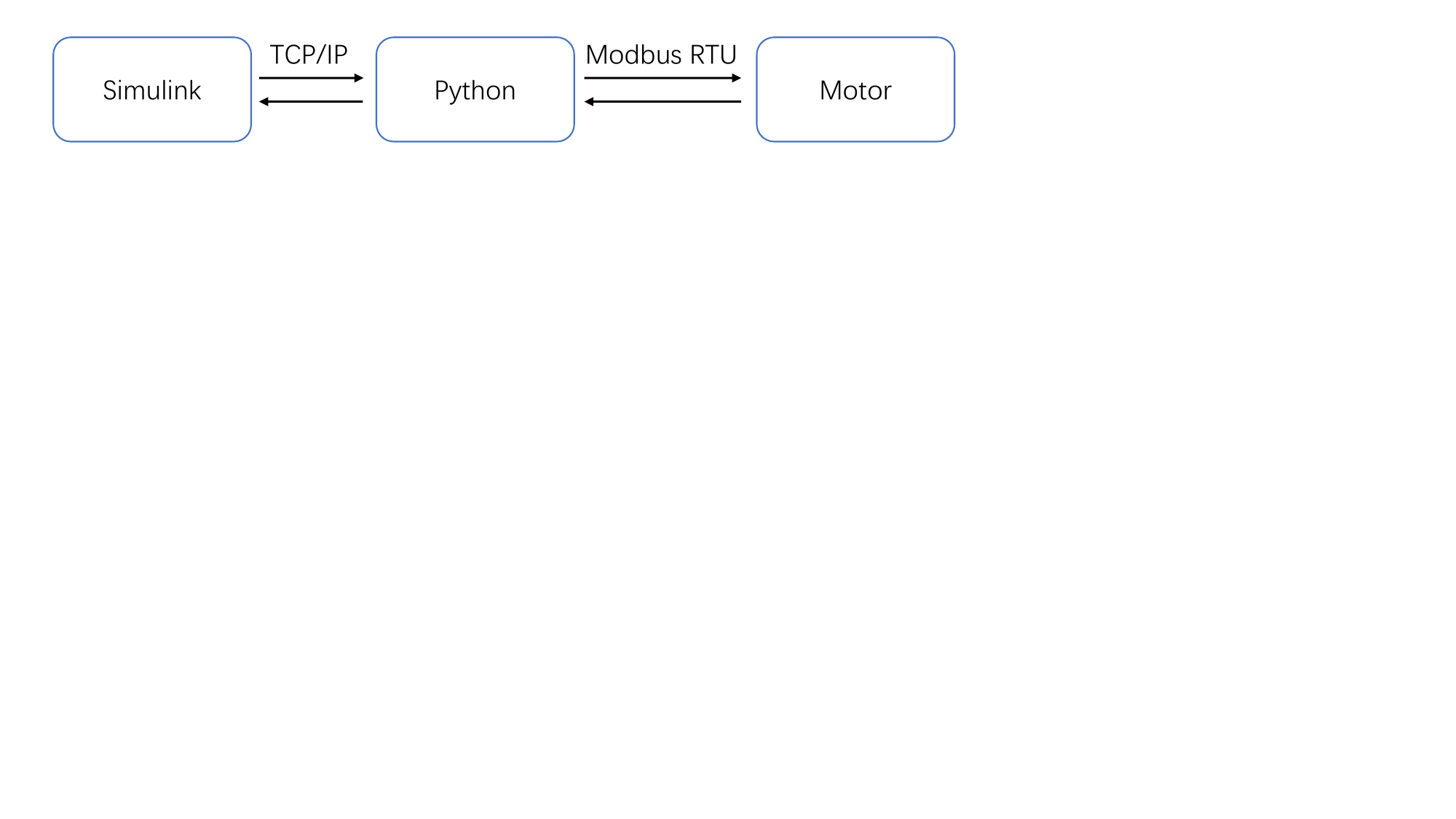}
	\vspace*{-5.6cm} 
	\caption{ Matlab Simulink and Python integrated structure via serial port wireless communication.}
	\label{robot_Communication}
\end{figure}

First, we established relatively simple sub-models for the controllers and regulators of the robots, all of which are designable. It is well known, to implement both model training in a simulated environment and real-time control in a real environment, modeling human-related variables and complex robot mechanics is a critical step. For modeling robot mechanics, we translated our SolidWorks Ankle robot model into a Simulink model (see Figure \ref{robot_simulink}) and conducted both RL training and real-time testing in a Matlab Simulink and Python integrated structure via serial port wireless communication (e.g., TCP/IP) (see Figure \ref{robot_Communication} for the communication between high-level RL training and low-level motor control).

Modeling human behavior is always challenging due to the nonlinear, time-varying nature of human nervous systems. In this study, to model the human agent without detailing the complexities of the human cognitive system, we treat the human as an input-output system. The inputs are various stimuli, including sound, light, electrical signals, and complex interfaces like Virtual Reality (VR) and/or Augmented Reality (AR), which humans can perceive and respond to. The outputs are the human reactions (biofeedback) to these stimuli, which we cam measure using sensors. For example, in our previous research on audio-stimulated bicycle cardiovascular exercise (see \cite{argha2017heart}), we used an \textbf{input-output-model-based} approach to regulate heart rate response along a predetermined reference profile during cycle-ergometer exercises. We designed a system that transmitted commands as auditory biofeedback to adjust exercise intensity via the patient biofeedback system.  Specifically, this method ensures that exercise intensity is effectively modulated by a beep generated by a computer audio system in response to real-time heart rate measurements. 
 By applying input-output modeling and multiple model strategy, we can create multiple sub-models of patient biofeedback, allowing us to individualize the types of biofeedback.

\begin{figure}[h]
	\centering
	\vspace*{-0.2cm}
	\hspace*{-0.0cm}\includegraphics[width=0.05\textwidth]{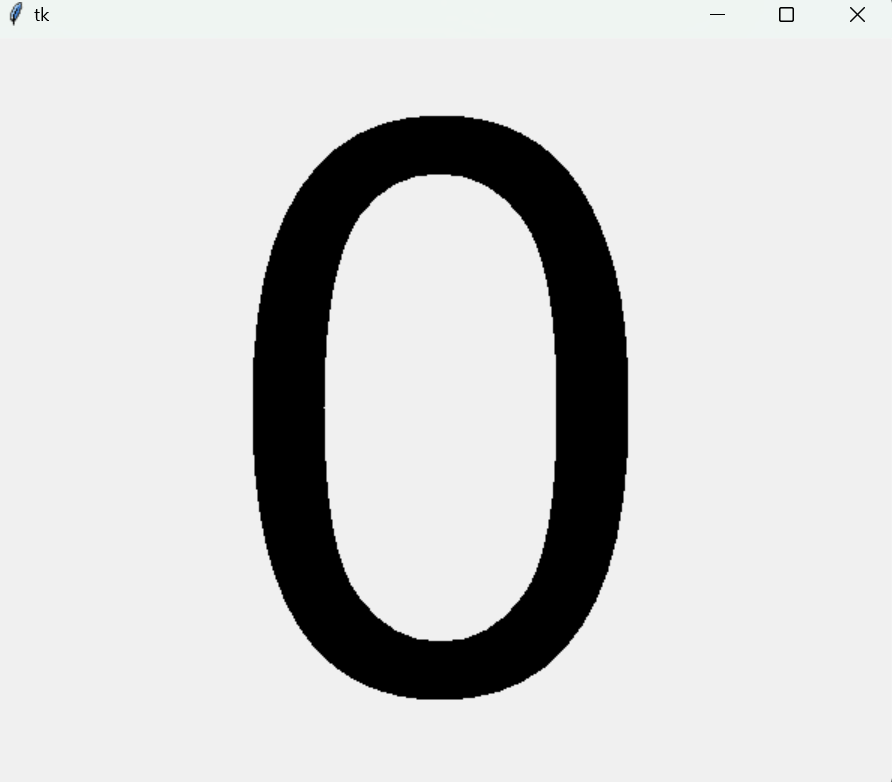}
	\hspace*{-0.0cm}\includegraphics[width=0.05\textwidth]{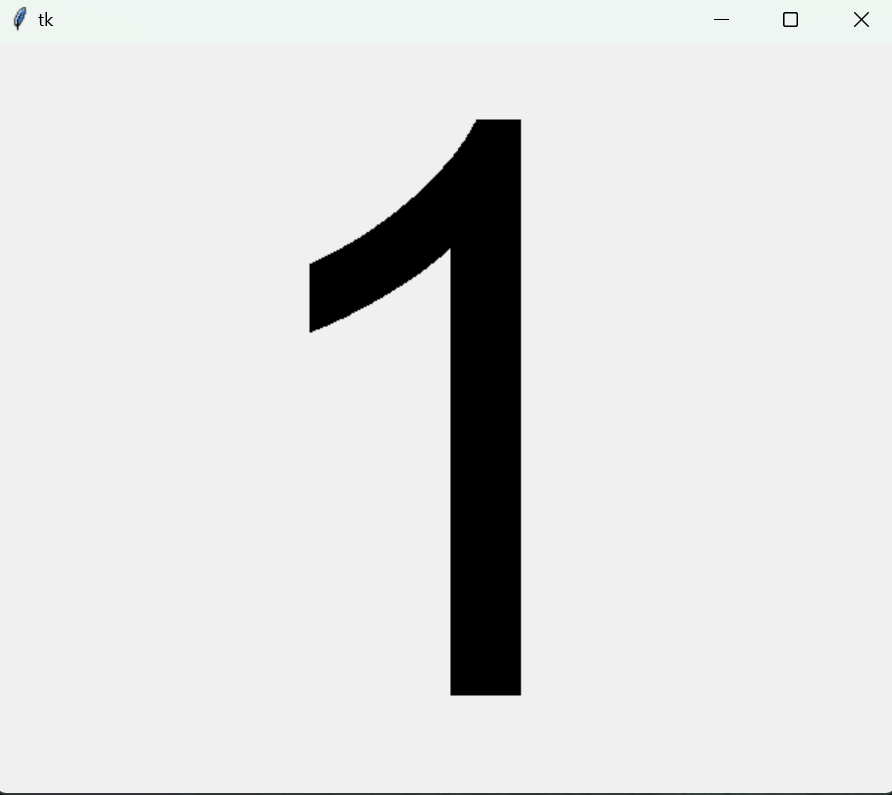}	
	\hspace*{-0.0cm}\includegraphics[width=0.05\textwidth]{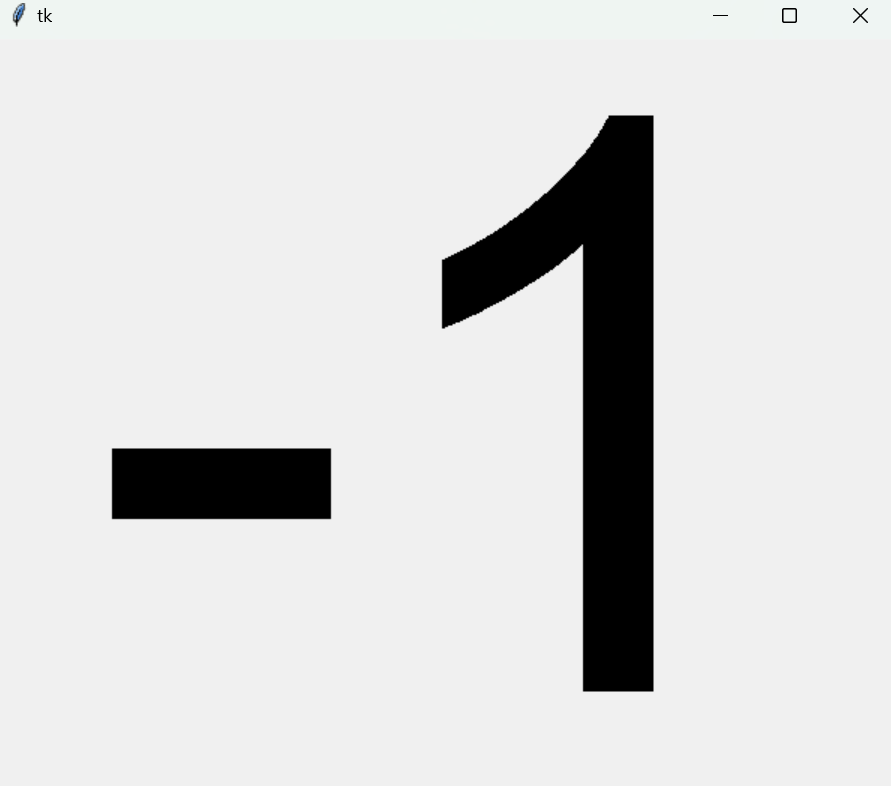}
	\hspace*{-0.0cm}\includegraphics[width=0.05\textwidth]{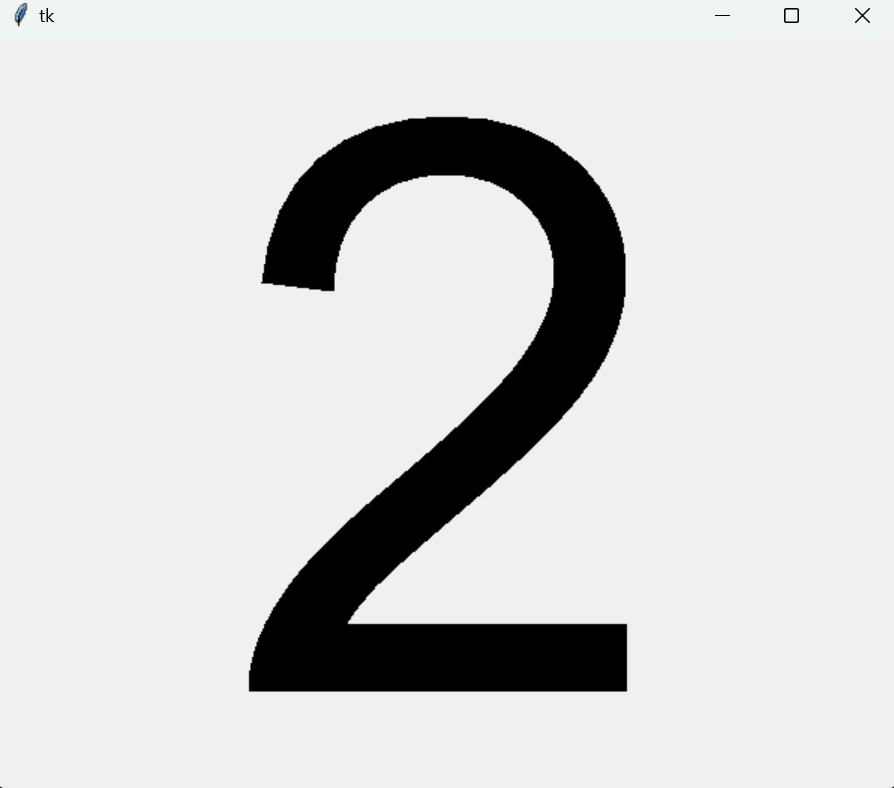}
	\hspace*{-0.0cm}\includegraphics[width=0.05\textwidth]{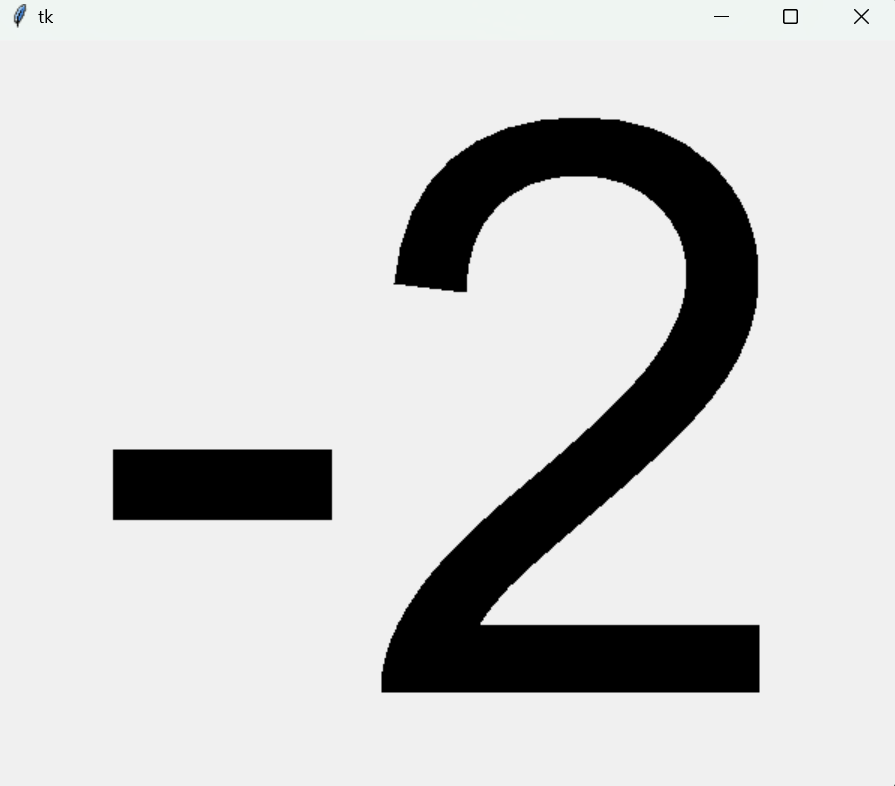}
	\vspace*{3.5cm} 
	\vspace*{-3.5cm} 
	\vspace*{0cm} 
	\hspace*{-0.6cm}\includegraphics[width=0.45\textwidth]{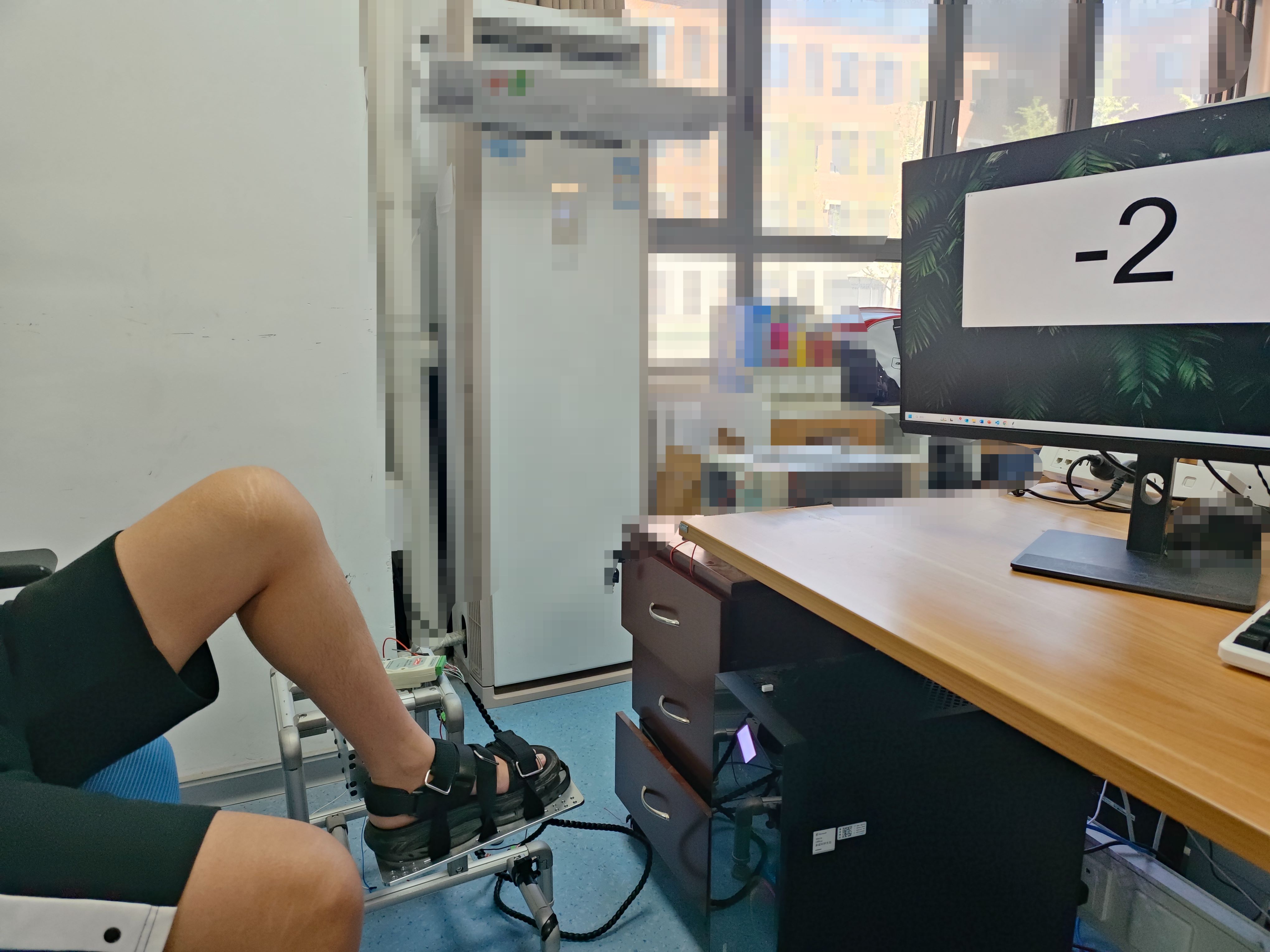}\\
	\vspace*{0.cm} 
	\caption{Visual stimulation in Robot-Assisted Rehabilitation}
	\label{fig:shared_states}
\end{figure}

\begin{figure}
	\centering
	\includegraphics[width=0.3\textwidth]{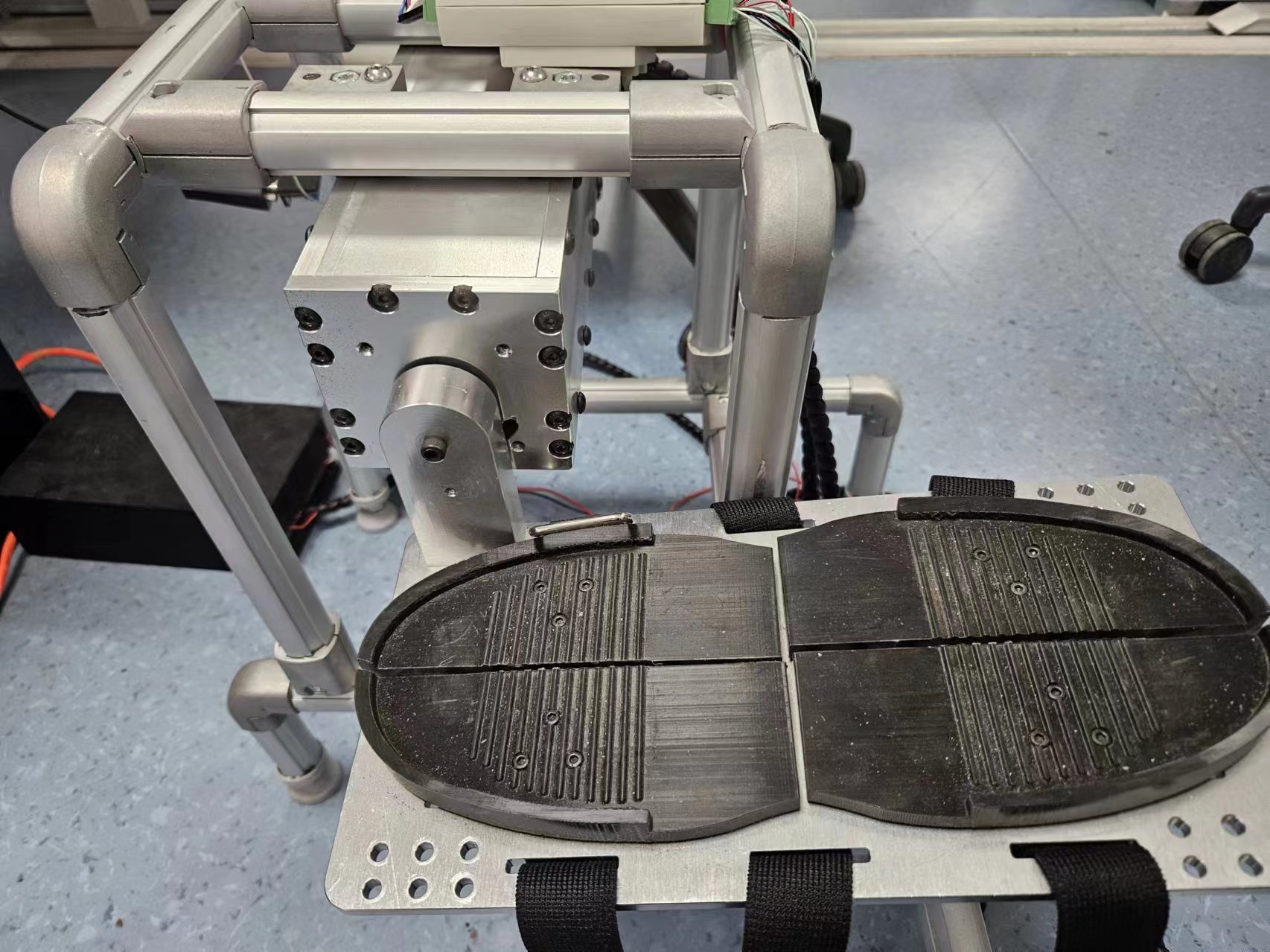}
	\caption{Ankle rehabilitation robot developed at the Jinan Key Lab of Intelligent Rehabilitation Robotics.}
	\label{fig_Robot}
\end{figure}

For this study, we developed a visual indicator in a set of digits $\{0, 1, -1, 2, -2\}$ (see Figure \ref{fig:shared_states}) to indirectly train the cognitive system using visual stimuli to influence the torque generated by the patient's lower limb musculoskeletal system. This torque can be measured using pressure sensors embedded in the foot pedal, as shown in Figure \ref{fig_Robot}. The patient can apply pressure to the foot pedal using either the toes or the heel, thereby affecting the front-end and back-end pressure sensors, respectively. We can build sub-input-output biofeedback models (i.e., visual stimuli as the input and the differential readings of the front and back pressure sensors) according to different rehabilitation stages and exercise locations, among other factors.

By treating the selection of multiple patient biofeedback models and robot sub-controllers as a co-adaptation process between two agents, we can implement a finite action space for both agents. This proposed multiple-model RL (MMRL) based adaptation is notably simpler than fully model-based or model-free RL approaches, streamlining the adaptation process.

In this study, to address the application of an ankle rehabilitation robot,  the multi-model strategy combined parameterized approaches during the RL training and testing procedures.  Specifically, we use neural networks to describe the adaptive sub-model switching policy, accommodating the complex relationship between states (observations) and actions in the rehabilitation process. This approach allows us to implement rehabilitation exercises efficiently without knowing the exact model structure of the sub-model selection logic. Instead, we focus on training the parameters (i.e., the weights of the neural network) to achieve better co-adaptation between human and machine based on dual-agent sub-model matching through optimal switching.


As a medical-related application, robot-based rehabilitation has unique characteristics \cite{blewitt2008millenson} \cite{candra2015investigation}. To ensure interpretability, transparency, and avoid overfitting, it is crucial to control model complexity. In the future, we may explore model reduction techniques, such as pruning \cite{maadi2021review} \cite{wang2022pruning}, to simplify the complexity of the neural network for the switching logic. This will help reduce model complexity and improve explainability, thereby enhancing acceptance among rehabilitation medical staff and patients.

The principal contributions of this study can be summarized as follows:

\begin{enumerate}
	\item A novel application of AR/VR technology is integrated into rehabilitation exercises, enabling adjustable difficulty levels tailored to the capabilities of stroke patients at various recovery stages.
	
	\item A multi-model input-output system is employed for the reinforcement learning (RL) framework, facilitating the optimization and ranking of sub-controllers according to value functions. This allows for a sophisticated analysis and decision-making process in real-time during rehabilitation sessions.
	
	\item The value function is innovatively utilized to modulate the extent of active or passive engagement required from patients during rehabilitation, enhancing the adaptability of the exercises to individual patient needs.
	
	\item A co-adaptive control strategy is implemented via RL, which dynamically alternates between fixed human input-output submodels and sub-controller models. This strategy enables personalized calibration of the rehabilitation process, determining optimal controller settings or ranking various human model inputs effectively.
	
	\item The study extends the design of PID sub-controller models and an assortment of human input-output submodels. These designs are rigorously evaluated through experimental trials, providing empirical evidence to support the efficacy of the proposed framework.
	
	\item This study contributes to the field by introducing a multiple model strategy that significantly reduces the action space, effectively simplifying the complexity of the actor and critic networks for both human and machine agents. Furthermore, the innovative design of the sub-models for each agent incorporates human expert knowledge, enabling a 'human-in-the-loop' approach. This advancement not only streamlines the decision-making process but also ensures that machine actions are closely aligned with human expertise and intuition, enhancing the overall effectiveness and applicability of the system.
	
\end{enumerate}

Overall, this research introduces an advanced dual-agent, multi-model reinforcement learning paradigm for robot-assisted ankle rehabilitation, promising significant strides in personalized, adaptive therapeutic interventions.

\section{Settings of Simulation training Environment and real-time testing environment}

In this study, the primary rehabilitation device utilized is an ankle rehabilitation robot, as depicted in Figure \ref{fig:shared_states} and \ref{fig_Robot}. Developed at the Jinan Key Lab of Intelligent Rehabilitation Robotics, this robot's pedal is equipped with force sensors that measure the pressure exerted by the toes and heels of the rehabilitation exerciser (Agent$_0$). These measurements are crucial for designing a controller that drives the pedal based on the differential readings of the front and back pressure sensors, as discussed in the introduction section. The difference in pressure readings from the front and back sensors is amplified by a sub-controller (often a PD controller) to control the pedal motors (for details about the sub-controller or regulators, see Table \ref{table:pid_models_rewards}).

On the other hand, the machine (Agent$_1$) provides a control output supplied by its sub-controller (often a PID controller).

For a robot-assisted rehabilitation system, RL training, testing, and analysis can be implemented in three distinct scenarios—real-world, simulated, and abstracted environments. In our previous research \cite{guo2024cooperative}, the convergence analysis of the CAMDP model was treated in an abstract environment, where state, action, and time are abstractions of real rehabilitation scenarios. In this study, we mainly investigate the real-world and simulated environments.

\begin{figure}[ht]
	\centering
	\vspace*{-0.3cm} 
	\hspace*{0cm}\includegraphics[width=0.45\textwidth]{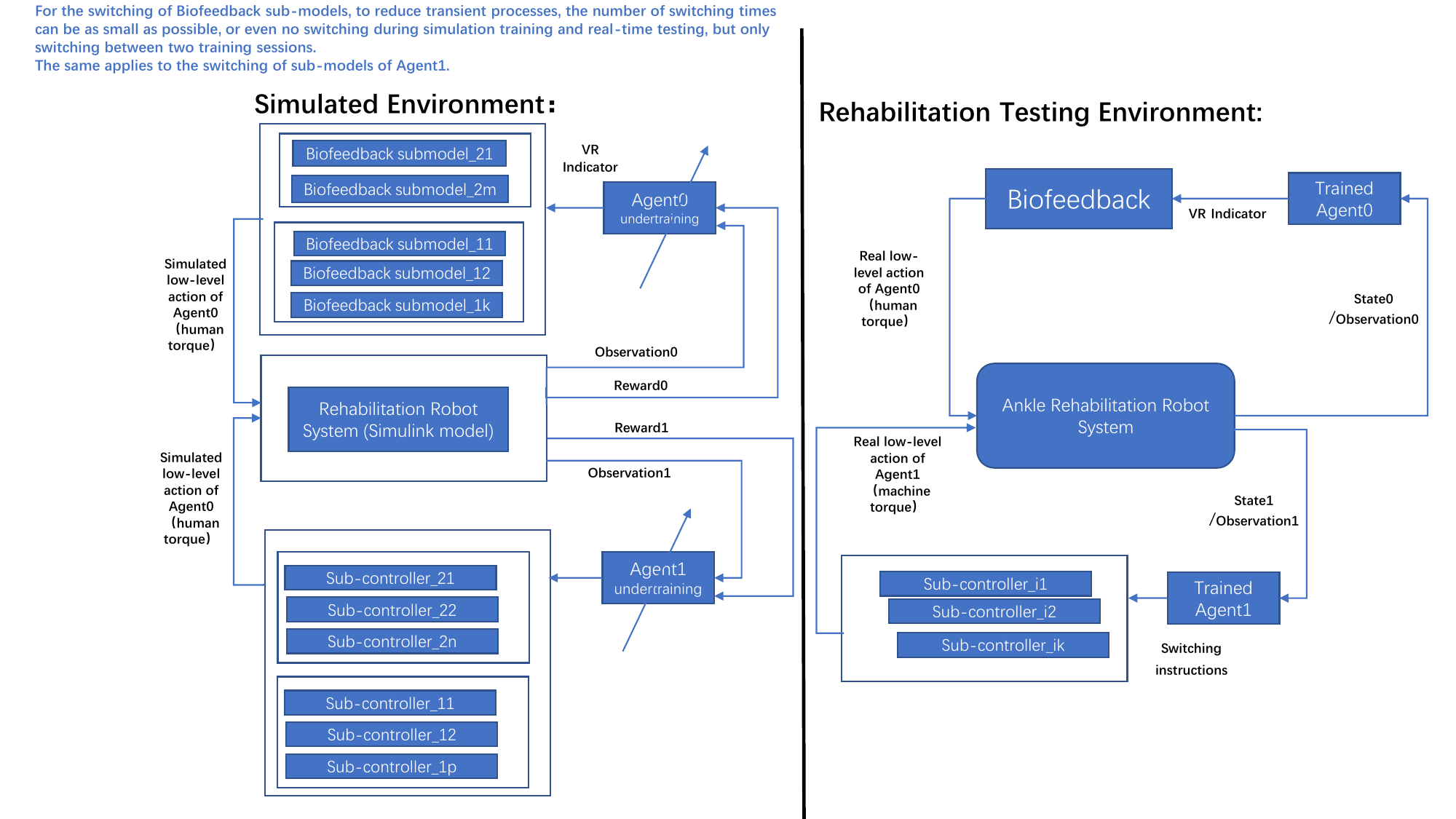}
	\vspace*{-0.4cm} 
	\caption{The block diagram of the training and testing environments.}
	\label{fig_simulink}
\end{figure}

We briefly introduced the real-world setting of the ankle rehabilitation system. To set up the simulated environment, as discussed in the introduction, we used Simulink/Matlab to model the machine (the rehabilitation robot). The input-output biofeedback model of the patient is also embedded in the Simulink/Matlab model. The simulated environment is used to establish the training environment for the switching networks (actor and critic networks). These trained networks are then applied in real-world experiments via TCP/IP serial port communication. Specifically, within simulated environments, the robot's model using Simulink/SimMechanics is shown in Figure \ref{robot_simulink}, and the overall structure for training and testing is depicted in Figure \ref{fig_simulink}. The input-output model representing human interaction, from the initiation of movement to the force applied, is also included in Figure \ref{fig_simulink}.

\subsection{The overall structure of the system}
\begin{figure}[ht]
	\centering
	\vspace*{-0.3cm} 
	\hspace*{-0cm}\includegraphics[width=0.45\textwidth]{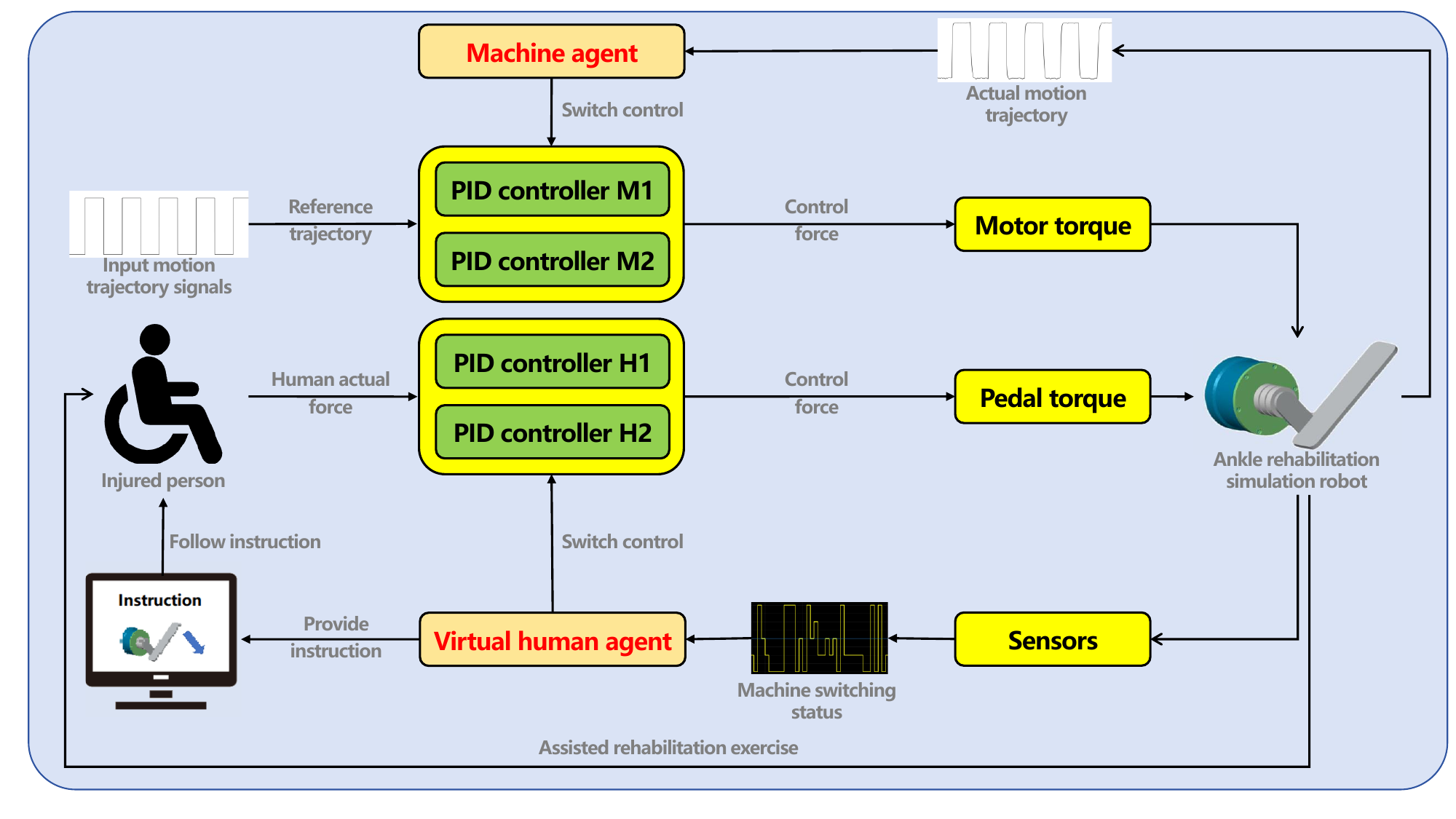}
	\vspace*{-0.4cm} 
	\caption{Block diagram of the multi-agent rehabilitation system.}
	\label{fig_Anfigure1}
\end{figure}

The proposed block diagram of the multi-agent rehabilitation system is shown in the Figure \ref{fig_simulink} and \ref{fig_Anfigure1}. In the initial phase, a machine agent and a virtual human agent are separately instantiated. The machine agent serves to govern the dual operating modes of the motor, each mode being associated with an individual sub-controller (here is a Proportional-Integral-Derivative (PID) controller). Similarly, the virtual human agent supervises the intensities of the patient's pedaling, with each intensity controlled by a PD controller. Subsequently, a motion trajectory signal is designated as the desired trajectory for the rehabilitation robot. Upon receiving the control mode from the machine agent, the corresponding PID controller is activated, subsequently regulating the ankle rehabilitation robot's motor torque to facilitate patients' rehabilitation training. Concurrently, the robot's sensors monitor the patient's exertion level, comfort, and modifications in motor mode in real-time, which serve as state/bservation variable inputted into the virtual human agent. The agent generates force guidance, which the patient follows displayed on the screen (the VR as shown in Figure \ref{fig:shared_states}). The actual pedaling intensity is dynamically adjusted by the sub-PD controller chosen by the virtual human agent, ensuring that the patient performs effective exercise within a comfortable setting.

\section{The Key Components of Dual Agent RL}

In this study, we will use dual-agent coadaptive control strategy based on reinforcement learning. In our previous study \cite{guo2024cooperative}, we proposed a CAMDP model to study the co-adaptation between patient and machine for the analysis of the convergence and switching behavior of the two agents. In \cite{guo2024cooperative}, we modified the finite MDP model based reinforcement learning to address the non-stationary phenomenon for MARL. Based on the framework in \cite{guo2024cooperative}, we can construct the similar structure of the two agents.

To this end, we should well define several key components of the model $M=<S_1, \, S_2, \, A_0,\, A_1, R>$, the state space (or observation space), action space and reword functions for both $Agent_0$ (the patient) and $Agent_1$ (the robot). It should be noted that the $Agent_0$ for the patient is a generalized concept, which includes all the variables can only be trained rather than designed. On the other hand, the varaibles which can be easily designed and manipulated belong to $Agent_1$, the intelligent robot or machine.

\begin{figure}[ht]
	\centering
	\vspace*{-0.3cm} 
	\hspace*{0cm}\includegraphics[width=0.40\textwidth]{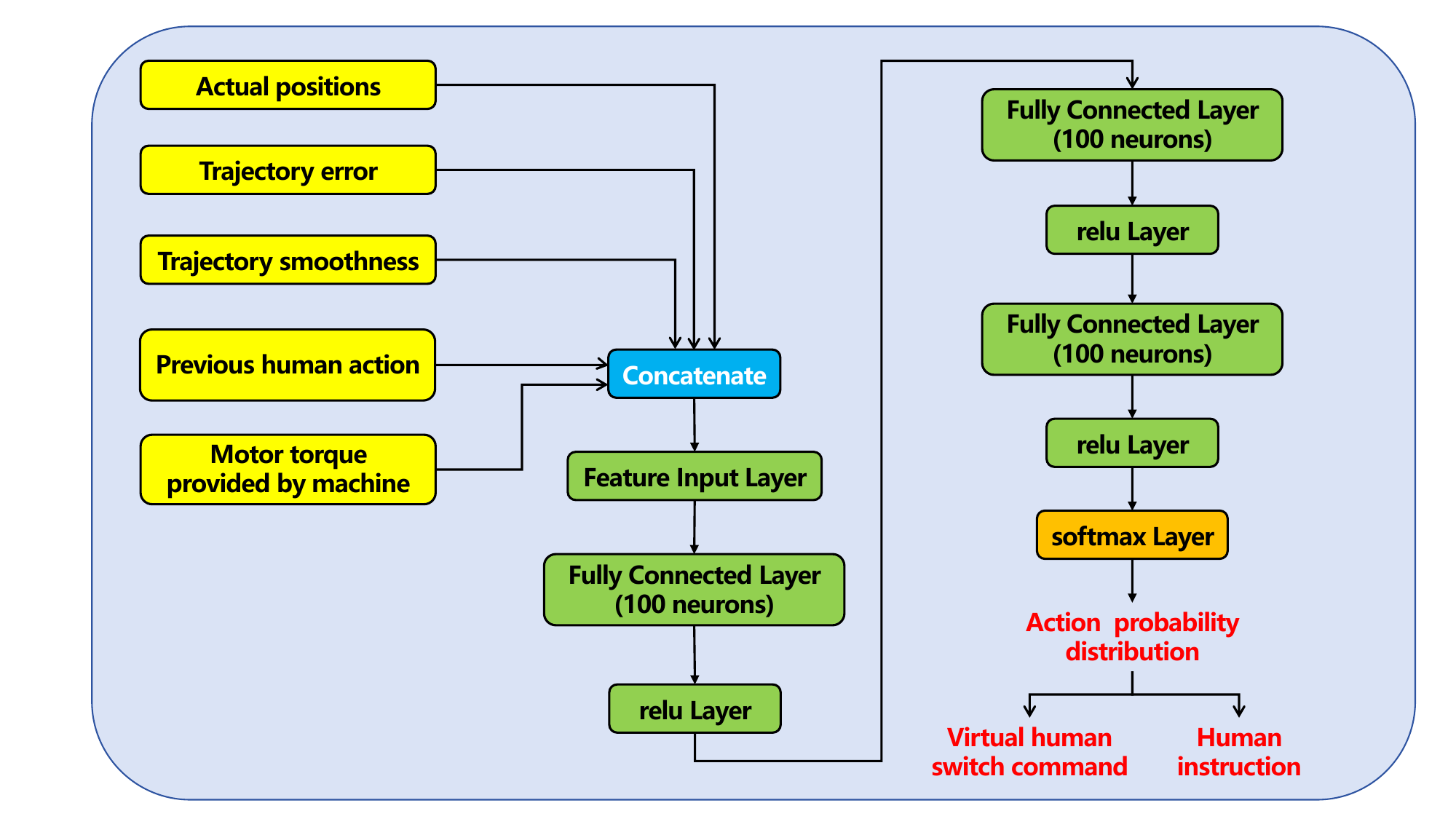}
	\vspace*{-0.2cm} 
	\caption{The actor network structure of the virtual human agent.}
	\label{fig_Anfigure3}
\end{figure}

Specifically, for Agent$_0$, in simulation, the state/observation includes simulated position ($p$), trajectory error ($e_t$), trajectory smoothness ($sm_t$), previous VR indicator value ($v_{\text{prev}}$), and motor torque provided by machine ($\tau_m$). The action is the VR indicator ($v_{\text{VR}}$) associated with different sub-PD controllers, represented by the policy function: $$v_{\text{VR}} = \pi_0(p, e_t, sm_t, v_{\text{prev}}, \tau_m).$$ 
In real-time testing, the state/observation are the same as the simulation setting, but the observations are all from the real world. The action remains the selected VR indicator (in a discrete range $\{-2, -1, 0, 1, 2\}$ as indicated by the HMI as shown in Fig.~\ref{fig:shared_states}) associated with different sub-PD controllers, but applied in the real environment (see Figure \ref{fig_Anfigure3}).

\begin{figure}[ht]
	\centering
	\vspace*{-0.3cm} 
	\hspace*{0cm}\includegraphics[width=0.55\textwidth]{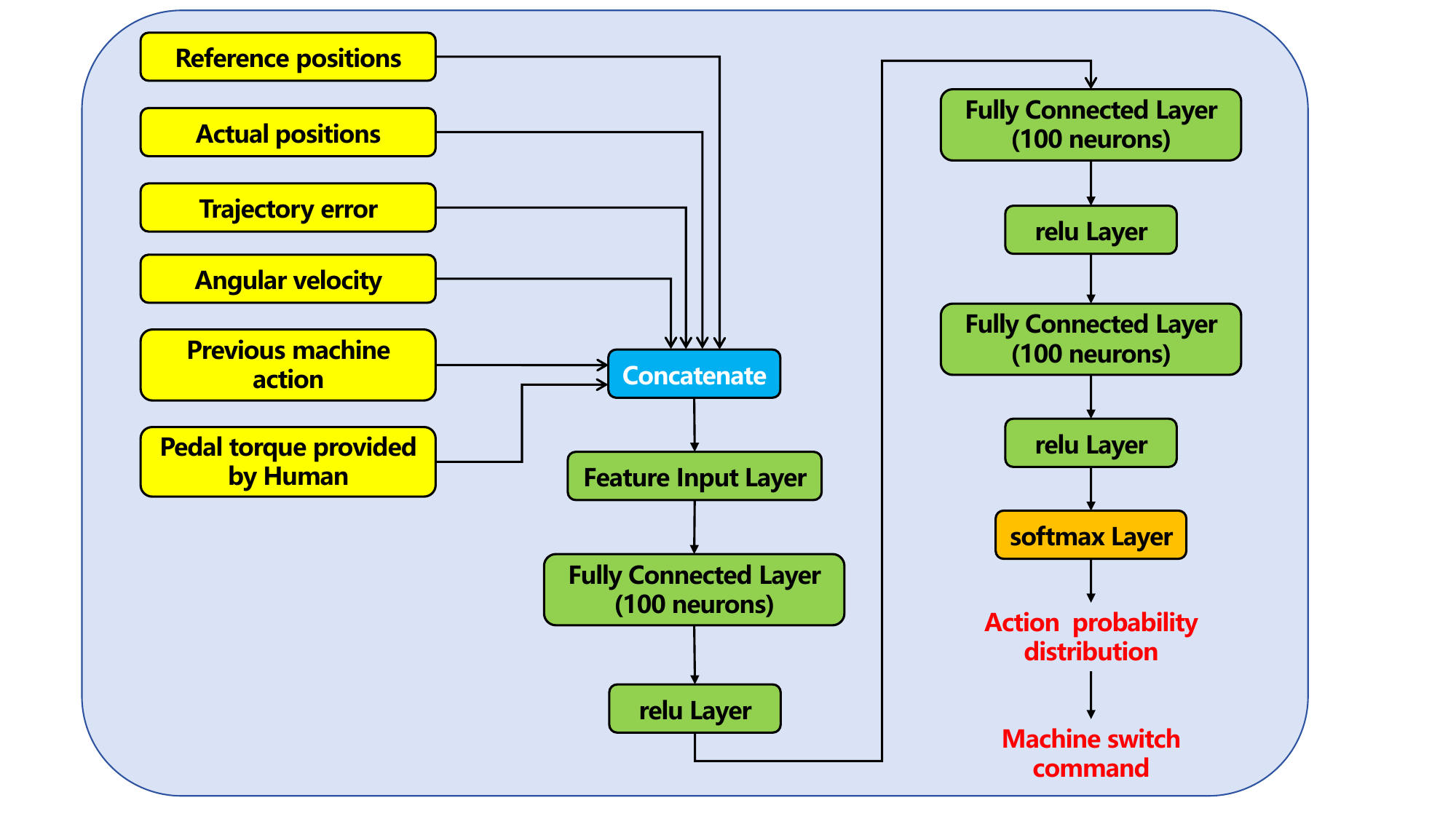}
	\vspace*{-0.7cm} 
	\caption{The actor network structure of the machine agent.}
	\label{fig_Anfigure2}
\end{figure}

For Agent$_1$, in simulation, the state/observation includes the reference positions ($r_p$), simulated motor position ($p_m$), trajectory error ($e_t$), angular velocity ($\omega$), previous machine action ($a_{\text{prev}}$), and torque provided by Human ($\tau_h$). The action is the selected sub-PD controller ($c_{\text{sub-PD}}$), represented by the policy function: $$c_{\text{sub-PD}} = \pi_1(r_p, p_m, e_t, \omega, a_{\text{prev}}, \tau_h).$$ 
In real-time testing, the state/observation are the same as the simulation setting, but the observations are all from the real world. The action remains the selected sub-PD controller, but applied in the real environment (see Figure \ref{fig_Anfigure2}).

\textbf{The reward function} generally can be structured into two main categories: continuous and enumerable variables. Continuous functions might include the error in tracking a predefined exercise trajectory; the patient's exercise effort quantified by the integrated torque or force they generate; and the assistive energy from the machine, measured by the power output of the machine. Enumerable variable could encompass the patient's emotional state determined through EEG or image-based emotion classification; and the physiological state of the patient estimated from wearable body sensors (e.g., heart rate monitors and wearable IMU sensors) and video cameras.

In this study, the reward function for the robot-assisted ankle rehabilitation system is treated as a continuous variable, encompassing both tracking error and the patient's exercise effort. Different weights can be assigned to these rewards depending on the rehabilitation stage and objectives. For instance, to highlight the patient's activity (i.e., emphasizing active rehabilitation), which is often crucial in the later stages of rehabilitation, we can allocate more weight to the patient's effort. Conversely, in the early stages of rehabilitation, to emphasize the machine's assistive role (i.e., emphasizing passive rehabilitation), we assign more weight to the tracking error. 

As this is a cooperative game, we combine the two agents' reward functions into one by using a weighted summation for both the human and the machine. For a detailed description of the reward function construction, see Subsection \ref{rewardfunc}.

This reward function setting is only a specific example. In fact, we can also select a combination of both continuous and categorical variables, especially for patient-related variables such as classified emotional states. Additionally, the weighting of the reward function is flexible and can be adjusted to suit different rehabilitation purposes.

\section{The Practical Structure of Multiple-Model RL}
In our previous study \cite{guo2024cooperative}, we proposed a Cooperative Adaptive Markov Decision Process (CAMDP) model to study the co-adaptation between patient and machine, focusing on the convergence and switching behavior of the two agents. 

While we established models for both agents in simulated environments, in practice, we adopted model-free reinforcement learning (MFRL) strategy due to the lack of an accurate mathematical model structure of the rehabilitation process. Model-free methods allow direct estimation of policy and value functions, adapting better to real-world variations and ensuring robustness.

To simplify the complexity of co-adaptation, we also introduced a multiple-model (MM) strategy, which focuses on optimizing the switching among different sub-models. Given the difficulty in identifying the switching logic structure, we applied neural networks (NNs) to implement this switching mechanism.

In this study, we adopted the MM strategy to leverage human experiences and simplify action spaces. This approach lies between MFRL and model-based RL (MBRL). While MFRL ignores model parameter estimation and MBRL requires exact model parameters, the MM strategy requires only a set of models to complete the RL or co-adaptive procedure.

It should be noted that even in a simulated environment with a model of the real world, model-free reinforcement learning (MFRL) is often used because exact modeling can be too complex or impractical. Model-free methods can generalize better to real-world variations and adapt to dynamics and changes not captured in the simulation, allowing for learning optimal policies directly from interactions.

Therefore, we combined the MM strategy with MFRL, using neural networks to construct the two agents. By optimizing the network weights, we implemented the co-adaptation of the two agents. For a detailed description of the network structures for both agents, see the following subsections.

\subsection{Network Structures of the Two Agents}

In both simulation and real environments, the virtual human agent and the machine agent each consist of an actor and a critic. The actor uses a three-layer neural network that takes states/observations as input and generates action probabilities. The machine agent manages the ankle rehabilitation robot's motor torque, ensuring precise motion along predetermined paths. It considers the pedal's reference position, real-time coordinates, rotational velocity, and trajectory deviations. To adapt interactively with the virtual human agent, it integrates previous actions and external forces from the patient. The actor network structure is shown in Figure \ref{fig_Anfigure2}.

The fundamental responsibility of the virtual human agent lies in guiding patients through rehabilitation exercises, aiming to maximize their repetitions while ensuring a comfortable environment. Consequently, the agent's state monitoring is intrinsically intertwined with the efficiency and continuity of the motion trajectory, serving as a vital indicator of effective guidance. It is imperative that the agent's observational parameters encompass the precise ankle position of the patient during these exercises, along with the deviation from the predefined reference trajectory for precision assessment. Despite the patient's diligent adherence to the agent's instructions, minor deviations may occasionally occur during the execution of rehabilitative actions. As such, it is essential to incorporate the real-time feedback of the patient's actual actions into the agent's state observation for continuous improvement and adaptation. Furthermore, to facilitate seamless human-machine collaboration, the inclusion of the machine's supportive torque and the patient's preceding actions within the state observation framework is indispensable. The actor network structure of the virtual human agent is shown in Figure \ref{fig_Anfigure3}.

\begin{figure}[ht]
	\centering
	\vspace*{-0.3cm} 
	\hspace*{-0.0cm}\includegraphics[width=0.55\textwidth]{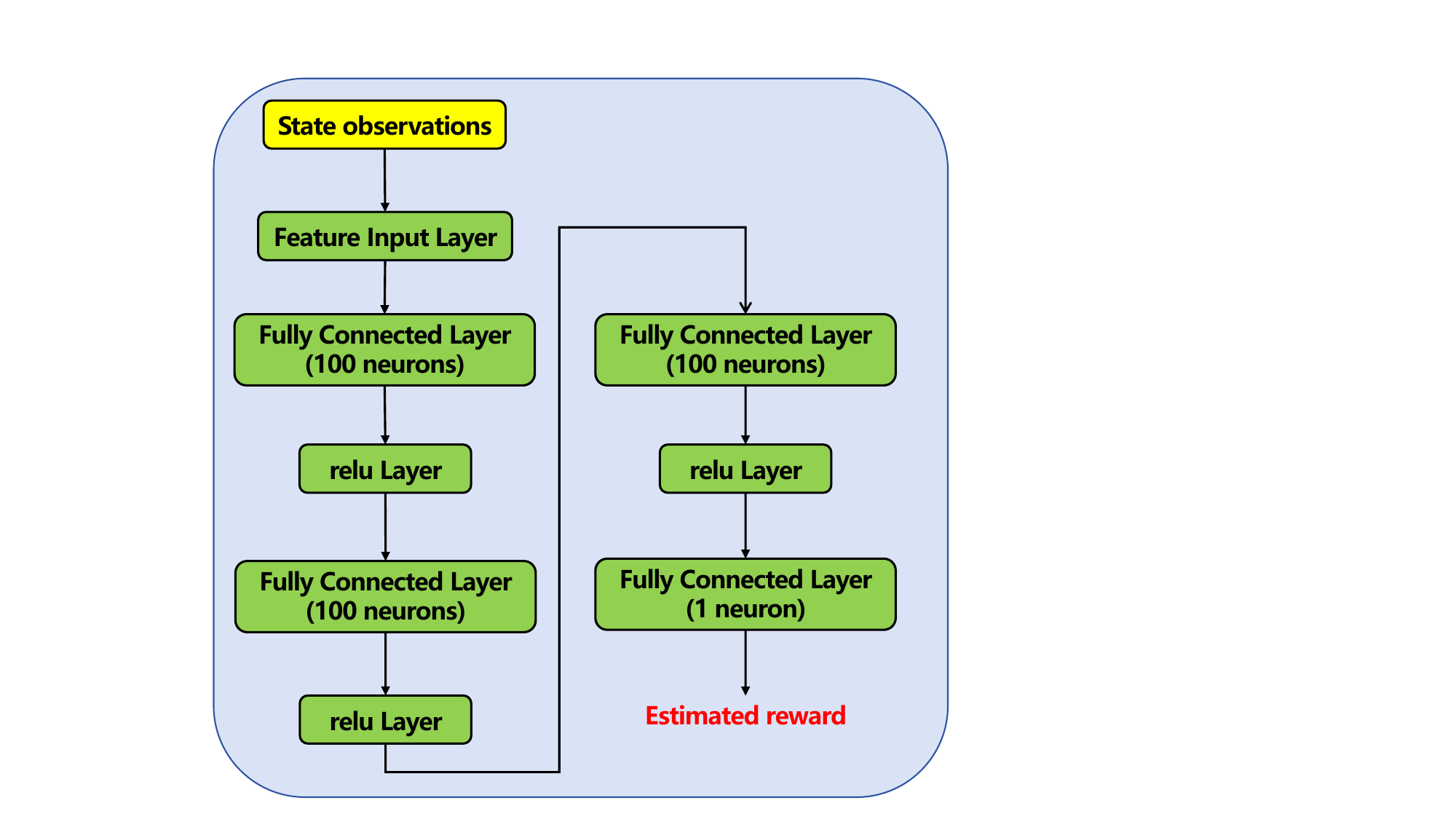}
	\vspace*{-0.6cm} 
	\caption{The critic network structure.}
	\label{fig_Anfigure4}
\end{figure}

In the proposed architecture, the evaluative framework for both the simulated human agent and its real counterpart adheres to a uniform structure. Both entities are modeled using a three-layer neural network architecture, which is primarily tasked with evaluating the value or desirability of a given state. The critic network functions by taking state information as input and computing an estimate of the expected cumulative reward, often referred to as the value function, following the execution of an action. The structure of the critic network is illustrated in Figure \ref{fig_Anfigure4}.

\subsection{Details about the construction of reward function}\label{rewardfunc} 
The construction of the reward mechanism is of paramount importance, as it directly influences the policy and behavioral decision-making of the virtual human agent during its interactions with the machine counterpart. The primary objective for the machine agent is to facilitate the patients' rehabilitation process through precise motion assistance. To achieve this, the machine's actual movement trajectory should closely align with the pre-established reference path, where precise movement is crucial for effective therapy. Therefore, the squared trajectory error is incorporated as a component of the reward function to penalize deviations from the desired path.

Additionally, maintaining an appropriate speed is crucial to prevent excessive stress and potential damage to both the machine and the patient. For instance, excessive speed can cause machine wear and tear or discomfort for the patient, making effective speed regulation essential. Thus, the reward scheme for the machine agent comprises two principal elements: position tracking reward and speed control reward. These elements ensure that the machine not only follows the desired trajectory but also does so at a safe and effective speed. A possible reward function for the machine  $r_{\text{machine}}$ can be defined as:
\begin{equation} \label{aeq1}
{r_{machine}}=\sigma\sum_{i=z-k}^{z}{(P_i-\widehat{P_i})}^2+\beta\ \omega_z
\end{equation}
where $k$ is the number of state observations; $z$ signifies the current time. $\sigma$, $\beta$ are weights of reward items; $P_i$ is the $i$-th pedal actual position; $\hat{P}_i$ is the $i$-th reference position; $\omega_z$ is the current pedal angular velocity. From the standpoint of the patient, the key objective is to guarantee a comprehensive ankle rehabilitation process and ensure safety and comfort during the device's utilization. The fundamental purpose of rehabilitation exercises lies in facilitating patients' activities by means of predefined movement pathways. Consequently, the virtual human agent's reward should also consider the precision in tracking the motion position and minimize the angular velocity, which can be the same as ${r_{machine}}$ (see equation (\ref{aeq1})). Only, simply set the motion tracking reward item $r_m$ as follows:
\begin{equation} \label{aeq2}
	{r_m}=\sum_{i=z-k+1}^{z}{(P_i-\widehat{P_i})}^2
\end{equation}
Furthermore, considering the inherent challenge of real-time patient feedback in rehabilitation, it is imperative for us to deduce their emotional state from the available data. In general, it is observed that motion trajectories characterized by uninterrupted movement with minimal acceleration, essentially near-zero acceleration, often evoke a sense of comfort among patients. The comfort-reward item $r_c$ is defined as follows:
\begin{equation} \label{aeq3}
{r_c}=\sum_{i=z-k+3}^{z}\left|P_i+P_{i-2}-2P_{i-1}\right|
\end{equation}
To attain the objective of active physical therapy, it is imperative for patients to engage proactively with the device by repetitively exerting force voluntarily to enhance recovery outcomes. Consequently, the virtual human agent plays a pivotal role in guiding patients towards frequent force application. We assess the reward of active movement through quantifying the volatility of actor's action. Despite the virtual human agent's capacity to offer guidance, potential discrepancies in patient compliance or reduced engagement necessitate the agent's adaptability to individual behavioral traits. To address these circumstances, our reward algorithm incorporates a regulatory component that acknowledges such deviations. In instances where patients deviate from instructions or demonstrate inadequate activation, the virtual human agent refrains from mandating increased exertions. Instead, the system dynamically adjusts to the patient's current capability, transitioning seamlessly from active to passive rehabilitation mode. The effort-reward function $r_e$ is defined as follows:
\begin{equation} \label{aeq4}
{r_e}=\frac{E}{k-2}\sum_{i=z-k+1}^{z-1}\left(A_i-\frac{1}{k}\sum_{i=z-k+1}^{z}A_i\right)^2
\end{equation}
where $A_i$ is the $i$-th action from the virtual human agent; $E$ is human effort coefficient; if the last action is unequal to the current action, $E$ is 1, otherwise, $E$ is 0. Ultimately, we amalgamate these rewards and modulate the significance of each by assigning appropriate weights. Consequently, the reward function $r_{\text{human}}$ for the virtual human agent is formulated as follows:
\begin{equation} \label{aeq5}
	{r_{human}}=\mu{{\ r}_m}+\kappa\ {r_c}+{{\rho\ r}_e}
\end{equation}
where $\mu, \kappa, \rho$ are weights of reward items.

Finally, if we do not punish the pedal angular velocity, considering the similarity of equations (\ref{aeq1}) and (\ref{aeq2}), as well as the components of (\ref{aeq5}), which align with the cooperative rehabilitation goal, we can, in practice, simply use equation (\ref{aeq5}) as the common reward function for both agents.

\section{Training the agents}
\begin{figure}[ht]
	\centering
	\vspace*{-0.3cm} 
	\hspace*{-0.1 cm}\includegraphics[width=0.35 \textwidth]{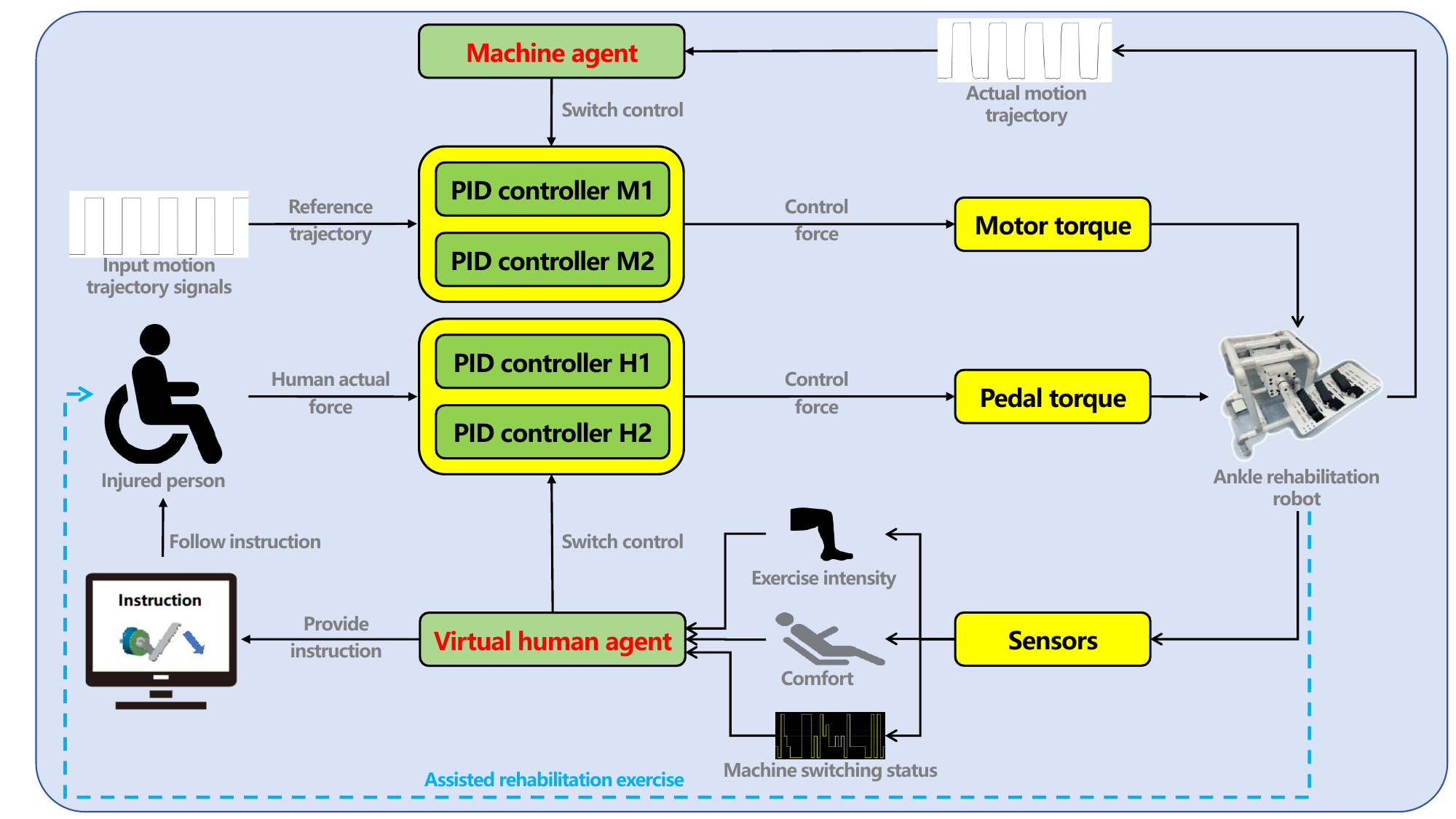}
	\vspace*{0cm} 
	\caption{The depicted schematic representation of training the multi-agent rehabilitation system.}
	\label{fig_Anfigure5}
\end{figure}
During the concurrent training of machine and virtual human agents, real patients are excluded to prevent early adaptation issues and control strategy failures. Offline model training ensures optimal performance without real-world interference. We use a virtual ankle joint model, assuming the virtual human's output mirrors expected patient behavior. Controlled random noise simulates performance discrepancies, ensuring system efficacy and practical utility. 

The depicted schematic representation of training the multi-agent rehabilitation system is presented in the Figure \ref{fig_Anfigure5}. The detailed training protocol proceeds as follows.

The training strategies for virtual agent intelligence and machine agent intelligence are identical. Initially, the actor engages in interactions with the environment to accumulate an extensive dataset, which is subsequently stored in the experience replay buffer. Given the current state \(S_t\), the actor computes the probability distribution of actions, from which a random action \(a_t\) is chosen. Following the execution of this action, the system transitions to the subsequent state \(S_{t+1}\), accompanied by a corresponding reward \(r_t\). The variable \(t\) signifies the current time. By calculating the return of each interaction, the conventional On-Policy strategy employs real-time rewards as proxies for long-term returns in gradient updates, often resulting in substantial algorithmic fluctuations. To mitigate these fluctuations, the Actor-Critic approach is adopted. This strategy involves utilizing a model to estimate the state values, thereby reducing the variance during optimization. Consequently, the advantage function \(Q_t\) is derived, leveraging the benefits of both policy evaluation and control in enhancing the learning process.

\[
Q_t = \sum_{i=t}^{N_e-1} \gamma^{i-t} (r_i + \gamma V(S_{i+1}, \phi) - V(S_i, \phi)) \tag{6}
\]

where \(N_e\) is the length of the experience buffer; \(\gamma\) is the discount factor; \(V(S_i, \phi)\) is the critic network with the parameter values \(\phi\) and the \(i\)-th state observation input. Then we can obtain the return \(R_t\):

\[
R_t = Q_t + V(S_t, \phi) \tag{7}
\]

During the update process, a batch of data is initially retrieved from the experience repository, where each datum is associated with the advantage function and cumulative return values. The critic's principal function is to forecast the overall score subsequent to taking an action in a given state. The predicted score for the subsequent time step should exhibit a deviation, equating to the immediate single-step reward garnered at the current time point. The critic network aims to minimize the discrepancy between the predicted cumulative return and the actual return, incorporating the advantage function to optimize the estimation accuracy and stabilize learning. Consequently, grounded in these theoretical foundations, the loss function \(L_c(\phi)\) of the critic network can be formally defined as follows:

\[
L_c(\phi) = \frac{1}{2B} \sum_{n=1}^{B} (R_n - V(S_n, \phi))^2 \tag{8}
\]

where \(B\) is the batch size. The principal component of the actor's loss is computed as the ratio between the novel policy and its preceding counterpart. To maintain training stability, a clipping mechanism is employed. First, a clipping threshold is set, establishing an upper bound at \(1+\varepsilon\) and a lower bound at \(1-\varepsilon\). Whenever the ratio between the updated actor and its initial state surpasses these bounds, the actor's loss is determined by the nearest predetermined constraint. This practice restricts the policy's adaptation to occur exclusively along the gradient vector confined within the established boundaries, preventing direct adjustments based on the raw ratio. This strategy effectively prevents drastic policy fluctuations over brief periods, thereby minimizing the likelihood of instabilities or oscillatory behavior during the learning process. The clipping mechanism enables the algorithm to approximate an optimal update at each iteration while preserving the practical constraints on policy revisions. Consequently, even if the newly proposed policy occasionally exhibits superior return prospects, it reverts to a secure boundary if it exceeds the allowable update parameters, thereby safeguarding the integrity of the current policy performance. Formally, the actor network's loss function \(L_a(\theta)\) can be mathematically expressed as:

\begin{align}
	L_a(\theta) = \frac{1}{B} \sum_{n=1}^{B} \Bigg(&-\min\left(\frac{T(a_n \mid S_n, \theta)}{T(a_n \mid S_n, \hat{\theta})} \cdot Q_n, \text{clip}_n(\theta) \cdot Q_n\right) \nonumber \\
	&+ \alpha K_n(\theta, S_n)\Bigg) \tag{9}
\end{align}

\[
\text{clip}_n(\theta) = \max\left(\min\left(\frac{T(a_n \mid S_n, \theta)}{T(a_n \mid S_n, \hat{\theta})}, 1+\varepsilon\right), 1-\varepsilon\right) \tag{10}
\]

where \(T(a_n \mid S_n, \theta)\) is the probability of taking the \(n\)-th action \(a_n\) when in state \(S_n\), given the updated policy parameters \(\theta\) in the batch set; \(\hat{\theta}\) is the previous policy parameters before current updating epoch; \(\varepsilon\) is the clip factor; \(\alpha\) is the entropy loss weight factor; \(P_n(\theta, S_n)\) is the entropy loss item. In the realm of reinforcement learning, an effective strategy necessitates not only the optimization of expected returns but also a judicious balance between exploitation and exploration to uncover rewarding actions. Entropy, as a quantitative indicator, gauges the inherent randomness or uncertainty within a policy, with higher values reflecting a greater propensity for exploration. The Proximal Policy Optimization (PPO) framework integrates a regularization mechanism, which promotes the maintenance of a stable exploration level concurrently with the objective of maximizing rewards. This strategy safeguards the policy against premature convergence towards suboptimal local optima. Additionally, entropy plays a crucial role in balancing immediate rewards with long-term exploration benefits. Overly narrow policies risk disregarding alternative, potentially more advantageous long-term scenarios. By augmenting entropy, PPO actively fosters policy diversity, thereby enhancing overall performance over extended periods. Moreover, high-entropy policies serve as a safeguard against overfitting, an issue that arises from excessive specificity in the policy. This allows the policy to exhibit enhanced generalizability across various environments. As a result, the entropy loss term, which embodies this balance between exploration and exploitation, assumes a pivotal role in the formulation of the PPO algorithm's objective function. Thus, the entropy loss item \(K_n(\theta, S_n)\) can be defined:

\[
K_n(\theta, S_n) = -\sum_{l=1}^{Y} T(a_l \mid S_n, \theta) \ln T(a_l \mid S_n, \theta) \tag{11}
\]

where \(Y\) is the number of actions. \(T(a_l \mid S_n, \theta)\) is the probability of taking the \(l\)-th action \(a_l\) when in state \(S_n\) following the current policy. In compliance with the established reward mechanism and model update procedure, the patient engages in interactive training with both the virtual human agent and the machine agent. Gradually, the patient acclimates to the mode transition rules of the rehabilitation robot during therapy, prompting the device to adapt its approach dynamically based on the patient's ongoing performance. This continuous exchange facilitates concurrent improvement for both parties, ultimately resulting in a synergistic adaptation between the participant and the rehabilitation device.

\section{The arrangements of the multiple models for both human and machine agents}

One of the major strategies of this study is the use of multiple model techniques. To effectively integrate this with the RL-based adaptation procedure, we first merged the multiple models into the action spaces of the two agents. This greatly reduced the complexity of the policy networks for both agents. Furthermore, we limited the size of the sub-controllers by carefully selecting suitable submodels to implement online model switching. This approach significantly minimizes the adverse effects of transient behavior due to model switching during both training and testing.

For the design of the sub-controllers and the sub-models, we used classical control design strategies, such as PD and PID control with different control parameters, to implement fine-tuning and co-adaptation. As we have built both simulation and real environments for training and testing, we can train the RL networks offline with different reward functions to serve various training purposes, such as fully passive rehabilitation exercise and semi-active rehabilitation, before real experimental testing. The details of the designed sub-controllers, reward functions, and achieved training results (value functions) are summarized in Table \ref{table:pid_models_rewards}.

\begin{table*}[h!]
	\centering
	\begin{tabular}{|c|c|c|c|c|c|c|}
		\hline
		\textbf{Setting} & \textbf{Reward Function} & \textbf{$\frac{Human\,subcontroller_1} {(PD \,\,controller)}$} & \textbf{$\frac{Human\,subcontroller_2} {(PD \,\,controller)}$} & \textbf{$\frac{Machine\,subcontroller_1} {(PID \,\,controller)}$} & \textbf{$\frac{Machine\,subcontroller_2} {(PID \,\,controller)}$} & \textbf{Value} \\
		\hline
		1 & $r_{m}+ r_{c}+ 5 r_{e}$ & 30, 0.2 & 15, 0.1 & 24, 2.4, 24 & 12, 1.2, 12 & 2124.89 \\
		\hline
		2 & $r_{m}+ r_{c}+ 5 r_{e}$ & 30, 0.2 & 15, 0.1 & 12, 1.2, 12 & 6, 0.6, 6 & 1940.50 \\
		\hline
		3 & $r_{m}+ r_{c}+ 5 r_{e}$ & 5, 0.1 & 2.5, 0.05 & 24, 2.4, 24 & 12, 1.2, 12 & 2617.20 \\
		\hline
		4 & $r_{m}+ r_{c}+ 5 r_{e}$ & 5, 0.1 & 2.5, 0.05 & 12, 1.2, 12 & 6, 0.6, 6 & 2738.69 \\
		\hline
		5 & $r_{m}+ 8 r_{c}+ r_{e}$ & 30, 0.2 & 15, 0.1 & 24, 2.4, 24 & 12, 1.2, 12 & 1272.84 \\
		\hline
		6 & $r_{m}+ 8 r_{c}+ r_{e}$ & 30, 0.2 & 15, 0.1 & 12, 1.2, 12 & 6, 0.6, 6 & 908.92 \\
		\hline
		7 & $r_{m}+ 8 r_{c}+ r_{e}$ & 5, 0.1 & 2.5, 0.05 & 12, 1.2, 12 & 6, 0.6, 6 & 460.14 \\
		\hline
		8 & $r_{m}+ 8 r_{c}+ r_{e}$ & 5, 0.1 & 2.5, 0.05 & 24, 2.4, 24 & 12, 1.2, 12 & 1626.40 \\
		\hline
	\end{tabular}
	\caption{Sub-Models and Training Rewards Summary}
	\label{table:pid_models_rewards}
\end{table*}

The table \ref{table:pid_models_rewards} provides a comprehensive summary of the designed sub-controllers, reward functions, and achieved training results (value functions). Each row in the table represents a different configuration of reward functions and control parameters for both human and machine sub-controllers. The configurations are numbered from 1 to 8 for easy reference.

\subsection{Reward Functions}
The reward functions used in the study are denoted by different combinations of \( r_{m} \), \( r_{c} \), and \( r_{e} \). These terms represent various components of the reward, such as machine effort, cooperative behavior, and error minimization, respectively. Two main types of reward functions are used:
\begin{enumerate}
	\item \( r_{m} + r_{c} + 5r_{e} \)
	\item \( r_{m} + 8r_{c} + r_{e} \)
\end{enumerate}
These reward functions are designed to balance the trade-offs between different aspects of the rehabilitation task.

\subsection{Sub-Controllers}
For each reward function, different sub-controller configurations are tested. The sub-controllers for the human agents are implemented using PD (Proportional-Derivative) controllers, while the machine agents use PID (Proportional-Integral-Derivative) controllers. The parameters for these controllers are given in the format:
\begin{itemize}
	\item PD Controller: \([K_p, K_d]\)
	\item PID Controller: \([K_p, K_i, K_d]\)
\end{itemize}

\subsection{Settings and Parameters}
\begin{itemize}
	\item \textbf{Setting 1}: Uses the reward function \( r_{m} + r_{c} + 5r_{e} \). The human sub-controller 1 is configured with parameters \([30, 0.2]\), and sub-controller 2 with \([15, 0.1]\). The machine sub-controller 1 uses \([24, 2.4, 24]\), and sub-controller 2 uses \([12, 1.2, 12]\). This Setting achieves a value function of 2124.89.
	\item \textbf{Setting 2}: Maintains the same reward function but with different machine sub-controller parameters: \([12, 1.2, 12]\) and \([6, 0.6, 6]\). The value function achieved is 1940.50.
	\item \textbf{Setting 3}: Keeps the reward function the same but changes the human sub-controller parameters to \([5, 0.1]\) and \([2.5, 0.05]\), achieving a value function of 2617.20.
	\item \textbf{Setting 4}: Similar to Setting 3 but with machine sub-controllers using \([12, 1.2, 12]\) and \([6, 0.6, 6]\), achieving a value function of 2738.69.
	\item \textbf{Setting 5}: Uses the reward function \( r_{m} + 8r_{c} + r_{e} \). The human sub-controllers use the same parameters as Setting 1, but the machine sub-controllers also use \([24, 2.4, 24]\) and \([12, 1.2, 12]\), resulting in a value function of 1272.84.
	\item \textbf{Setting 6}: Similar to Setting 2 but with the new reward function, achieving a value function of 908.92.
	\item \textbf{Setting 7}: Similar to Setting 3 but with the new reward function, achieving a value function of 460.14.
	\item \textbf{Setting 8}: Similar to Setting 4 but with the new reward function, achieving a value function of 1626.40.
\end{itemize}

\subsection{Analysis}
From the table, it is evident that different Settings of sub-controllers and reward functions lead to varied performance outcomes, as indicated by the value functions. Settings 3 and 4, using the reward function \( r_{m} + r_{c} + 5r_{e} \), tend to achieve higher value functions, suggesting better performance in terms of the defined reward components. On the other hand, Settings with the reward function \( r_{m} + 8r_{c} + r_{e} \) generally achieve lower value functions, indicating a potential trade-off between different aspects of the rehabilitation goals.

These findings highlight the importance of carefully selecting and tuning both the control parameters and reward functions to achieve optimal performance in cooperative rehabilitation tasks. The results demonstrate that by fine-tuning the controllers and appropriately defining the reward structure, it is possible to significantly enhance the efficiency and effectiveness of the rehabilitation process.

%
%

\section{The Training and Experimental Procedure}

This study implements reinforcement learning-supervised rehabilitation. The supervised rehabilitation process is divided into several steps for clarity:

\begin{procedure} \label{procedure__1}
	\item \textbf{Step 0:} Utilise Simulink robot models and pre-experimental data to identify a set of simple sub-models or controllers for various circumstances, along with input-output models for patient biofeedback involved in the feedback loop.
	\item \textbf{Step 1:} Employ the Actor-Critic network to train the two agents within the Matlab/Simulink environment, as illustrated in Figure \ref{fig_simulink}, until the training converges.
	\item \textbf{Step 2:} Conduct real-world experiments and use VR/AR (Human-Machine Interface (HMI)) to provide stimuli to the patients.
	\item \textbf{Step 3:} If the exercise effort does not meet the design specifications, adjust actions based on the experimental data to re-model the environment and redesign the controllers. This involves focusing on the modeling of the patient's biofeedback system, treated as an input-output system with the HMI indicator as the input and the response force as the output. Additionally, monitor the patient's condition using wearable sensor-based measurements, taking into account the patient's adaptation to the HMI indicator or the machine.
\end{procedure}

The overall system will be iteratively improved to achieve co-adaptation between the patient and the intelligent robot.

\section{Experiments}

%
%

According to the proposed procedure, we recruited 13 healthy young subjects, with details described in Table~\ref{tab:subject_details}. The experiment was approved by Shandong First Medical University under document number "R202402280067". The subjects were familiarised with the experimental procedure, which involved only the testing of the rehabilitation exercise, as the training of the four neural networks had been completed during simulation. The training results are shown in Table \ref{table:pid_models_rewards}.

Based on the training results, we followed Procedure \ref{procedure__1} and asked the subjects to perform rehabilitation under four different settings.

According to the pre-trained results shown in Table \ref{table:pid_models_rewards} for Subject 1, for the reward function $r_{m} + r_{c} + 5r_{e}$ (encouraging patients' active participation), we selected the sub-controller settings with the lowest value (Setting 2) and the highest value (Setting 4). For the reward function $r_{m} + 8r_{c} + r_{e}$ (encouraging robot assistance, resembling a passive rehabilitation mode), we selected the sub-controller settings with the lowest value (Setting 6) and the highest value (Setting 8) to complete the rehabilitation exercise.


\begin{table}[htbp]
	\centering
	\caption{Details of Recruited Subjects}
	\label{tab:subject_details}
	\begin{tabular}{|c|c|c|}
		\hline
		\textbf{Subject ID} & \textbf{Age} & \textbf{Gender} \\
		\hline
		1  & 30 & Male \\
		2 & 27 & Female \\
		3 & 36 & Female \\
		4 & 26 & Male \\
		5 & 56 & Male \\
		6 & 24 & Male \\
		7 & 33 & Male \\
		8 & 22 & Male \\
		9 & 29 & Female \\
		10 & 26 & Male \\
		11 & 25 & Female \\
		12 & 27 & Male \\
		13 & 26 & Male \\
		\hline
	\end{tabular}
\end{table}

\section{Results and discussions}

\begin{figure*}[h!]
	\centering
	\includegraphics[width=1\linewidth]{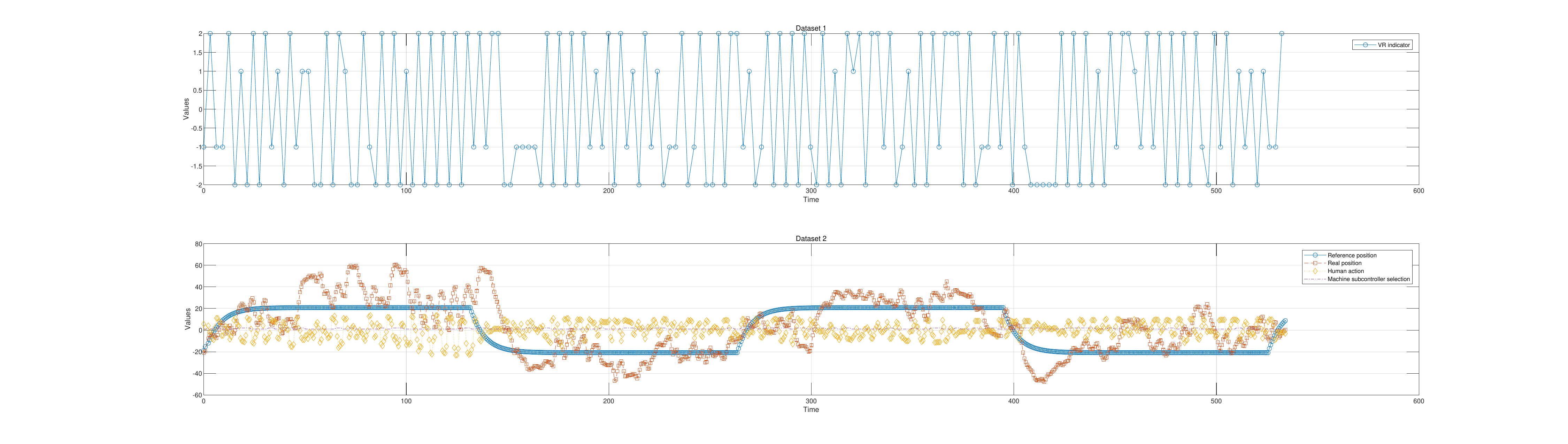}
	\caption{Comparison Plot for Subject 1 Setting 2.}
	\label{fig:anyang_2}
\end{figure*}

\begin{figure*}[h!]
	\centering
	\includegraphics[width=1\linewidth]{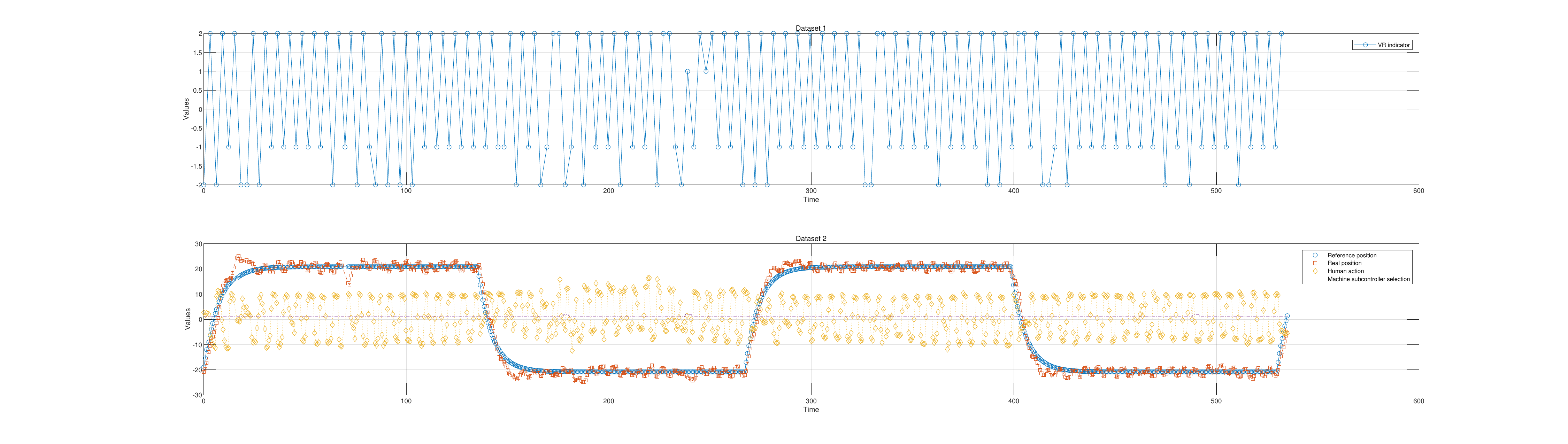}
	\caption{Comparison Plot for Subject 1 Setting 4.}
	\label{fig:anyang_4}
\end{figure*}

\begin{figure*}[h!]
	\centering
	\includegraphics[width=1\linewidth]{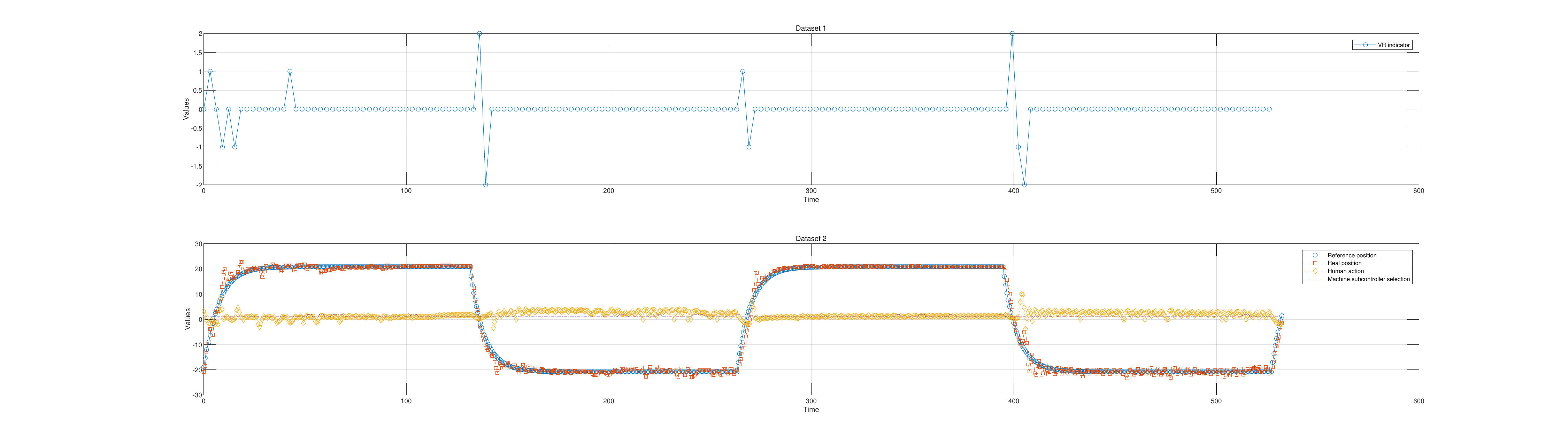}
	\caption{Comparison Plot for Subject 1 Setting 6.}
	\label{fig:anyang_6}
\end{figure*}

\begin{figure*}[h!]
	\centering
	\includegraphics[width=1\linewidth]{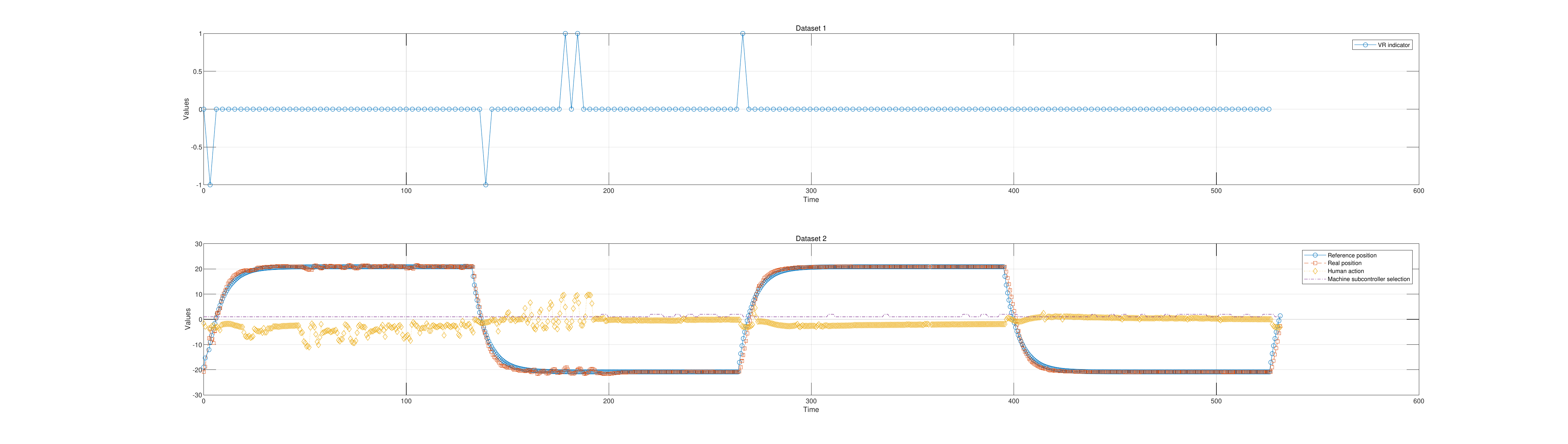}
	\caption{Comparison Plot for Subject 1 Setting 8.}
	\label{fig:anyang_8}
\end{figure*}

\begin{figure*}[h!]
	\centering
	\includegraphics[width=1\linewidth]{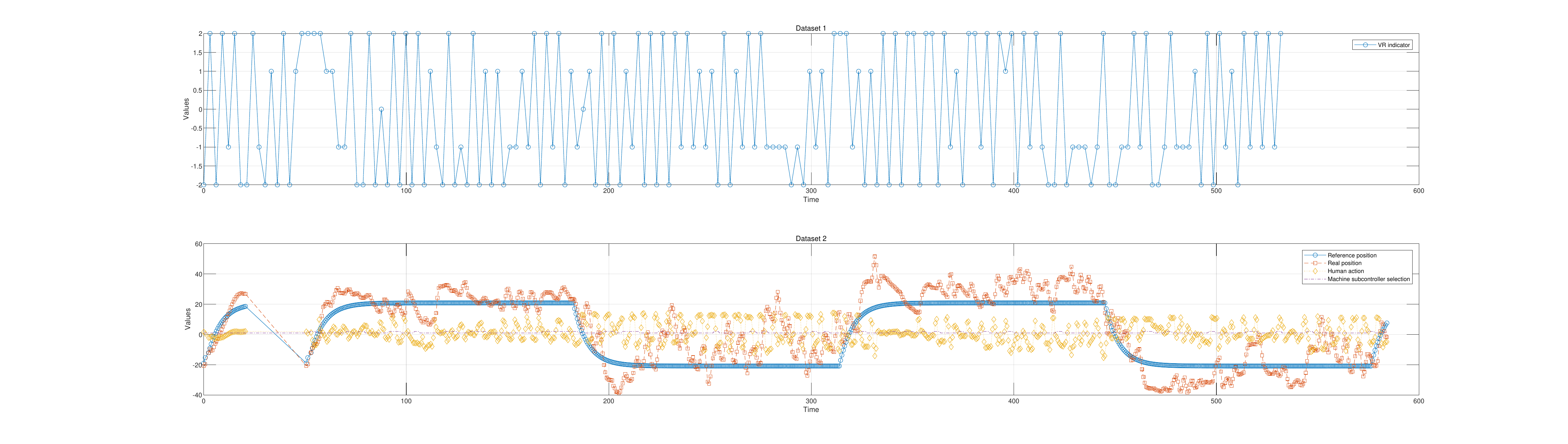}
	\caption{Comparison Plot for Subject 2 Setting 2.}
	\label{fig:huqifan_2}
\end{figure*}

\begin{figure*}[h!]
	\centering
	\includegraphics[width=1\linewidth]{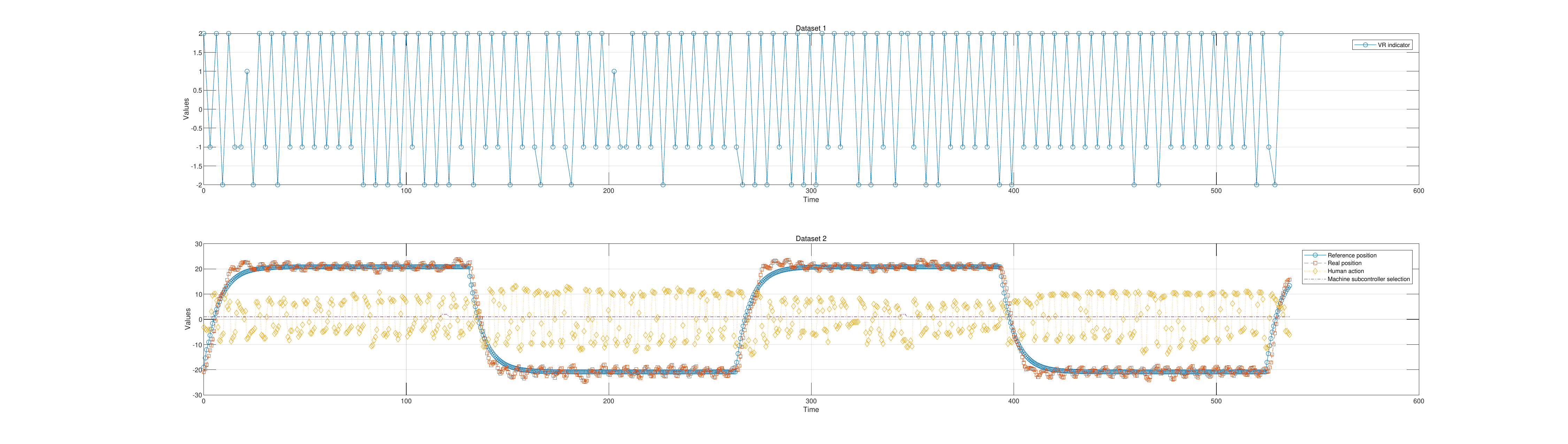}
	\caption{Comparison Plot for Subject 2 Setting 4.}
	\label{fig:huqifan_4}
\end{figure*}

\begin{figure*}[h!]
	\centering
	\includegraphics[width=1\linewidth]{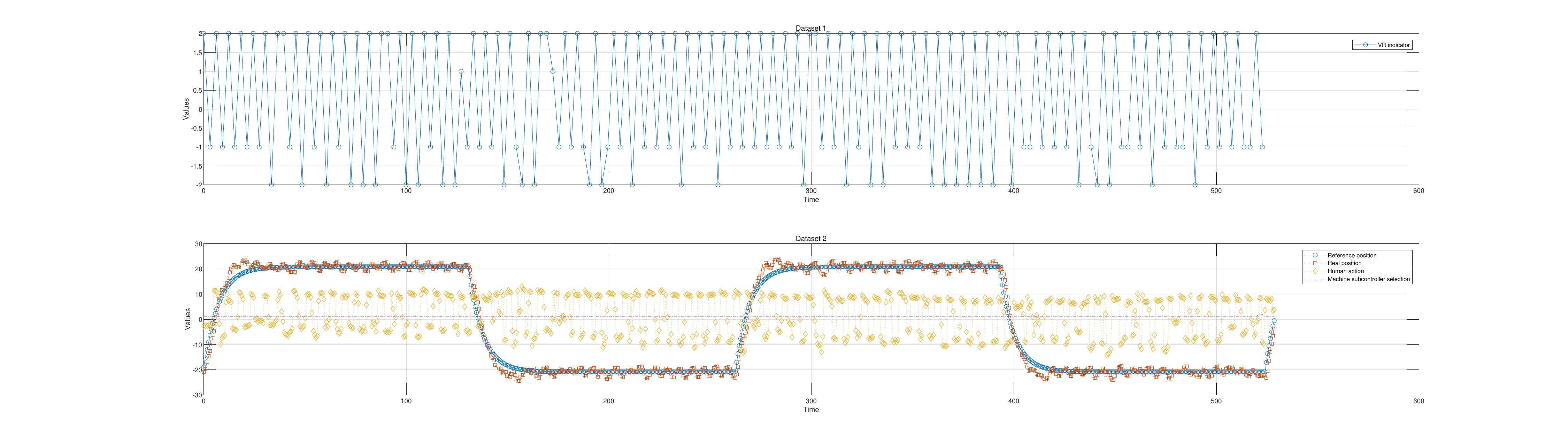}
	\caption{(Repeat of the Last Setting for Subject 1).}
	\label{fig:huqifan_4_2}
\end{figure*}

\begin{figure*}[h!]
	\centering
	\includegraphics[width=1\linewidth]{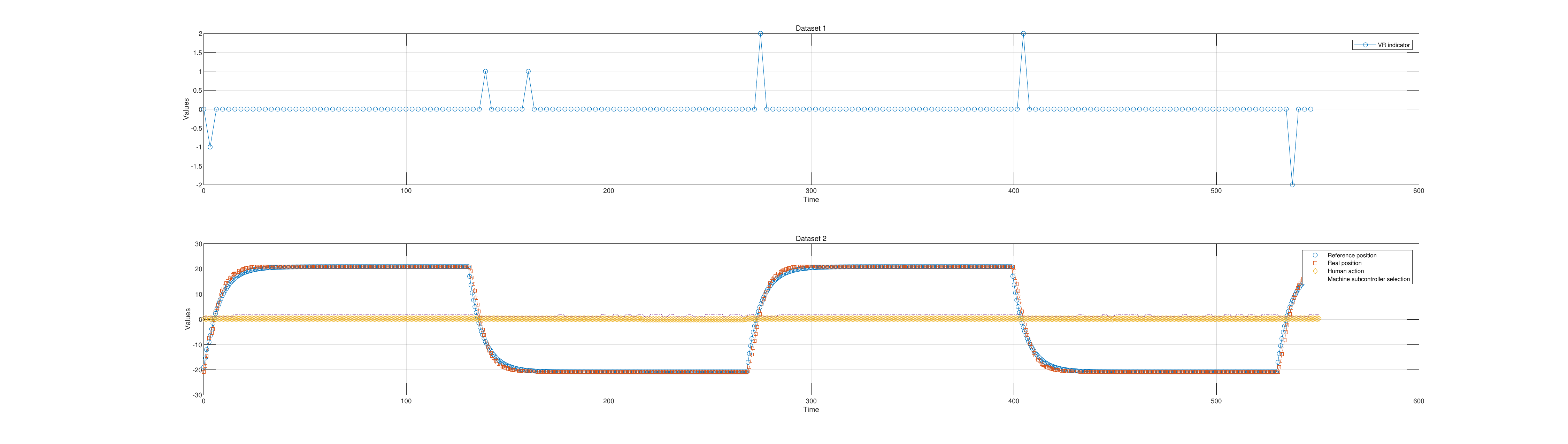}
	\caption{Comparison Plot for Subject 2 Setting 6.}
	\label{fig:wuren_1}
\end{figure*}

\begin{figure*}[h!]
	\centering
	\includegraphics[width=1\linewidth]{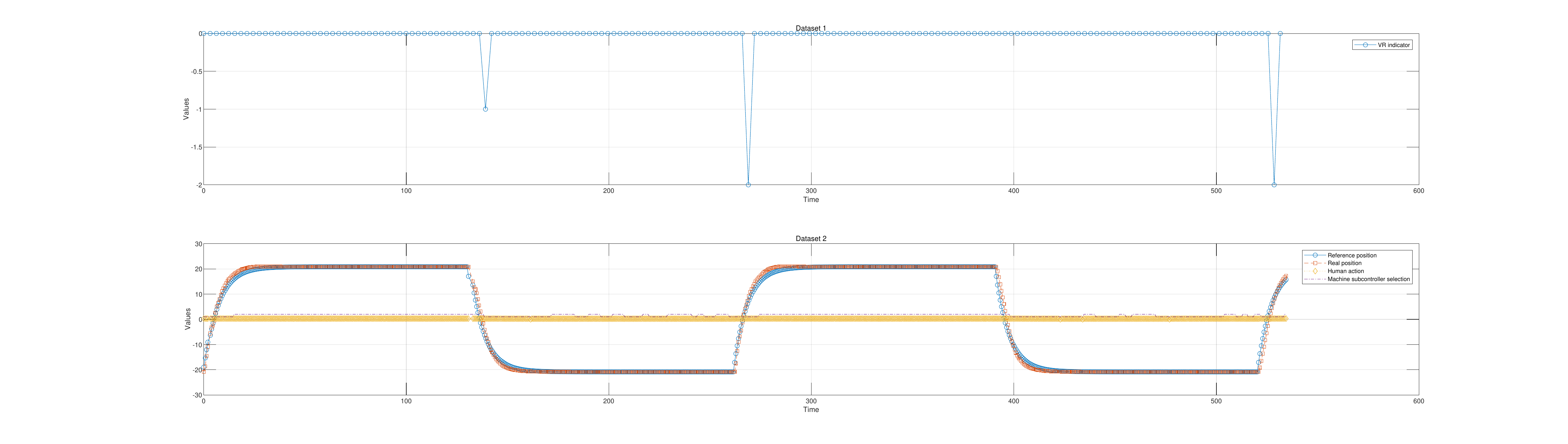}
	\caption{Comparison Plot for Subject 2 Setting 8.}
	\label{fig:wuren_2}
\end{figure*}

\begin{figure*}[h!]
	\centering
	\includegraphics[width=1\linewidth]{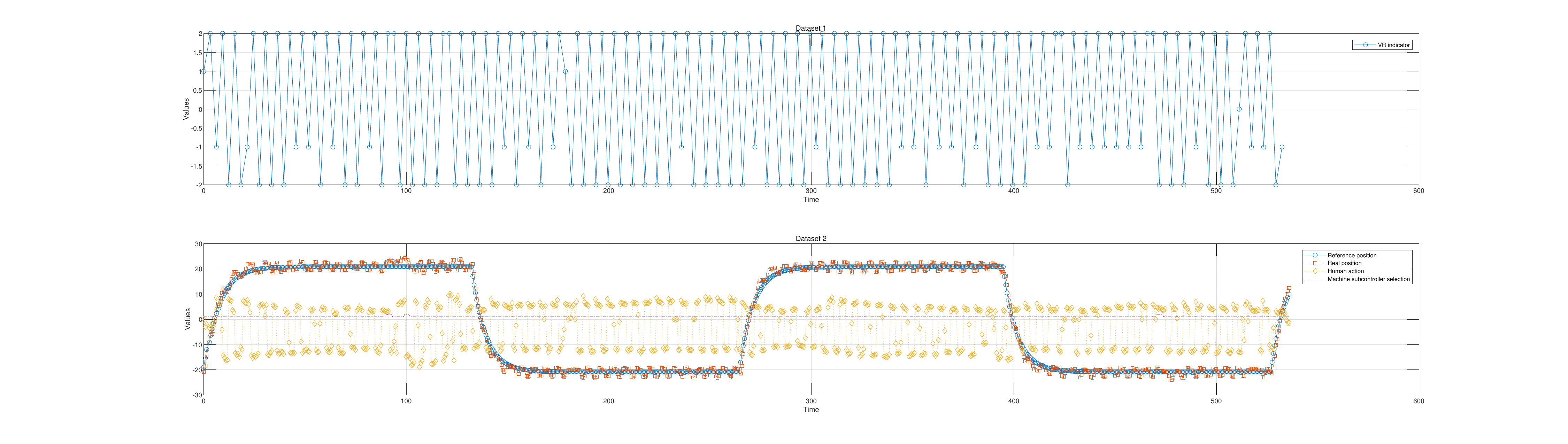}
	\caption{Comparison Plot for Subject 3 Setting 3.}
	\label{fig:liuhuifang_3}
\end{figure*}

\begin{figure*}[h!]
	\centering
	\includegraphics[width=1\linewidth]{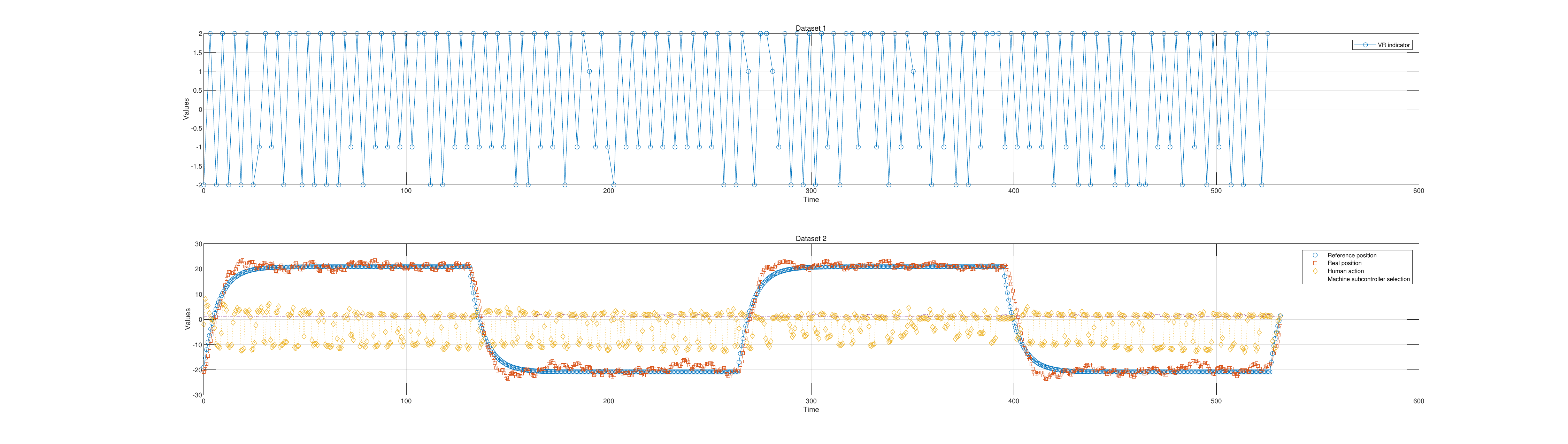}
	\caption{Comparison Plot for Subject 3 Setting 4.}
	\label{fig:liuhuifang_4}
\end{figure*}

\begin{figure*}[h!]
	\centering
	\includegraphics[width=1\linewidth]{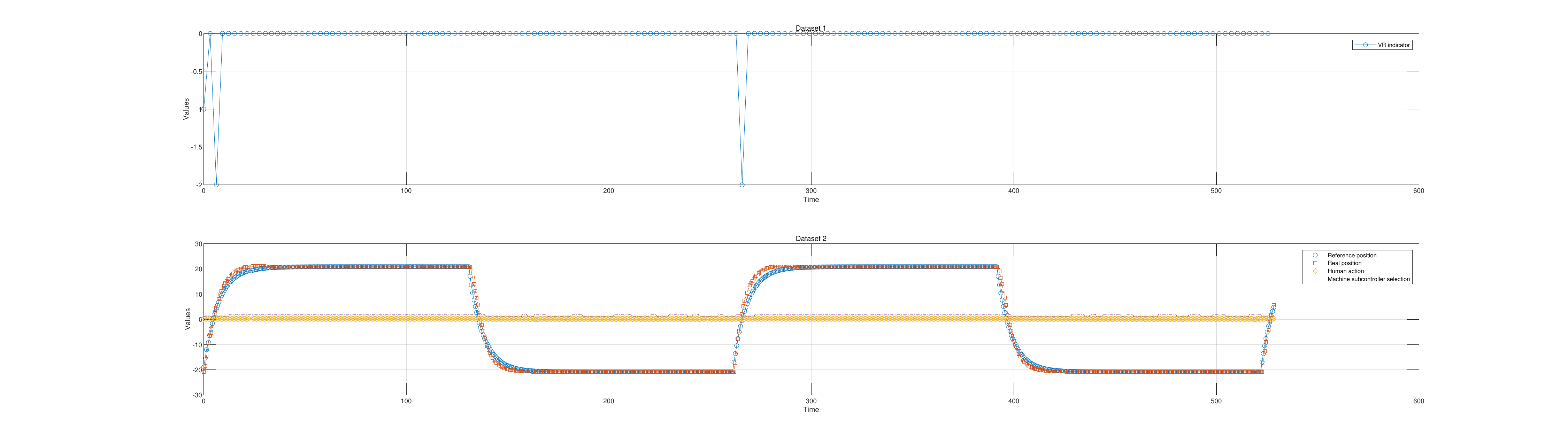}
	\caption{Comparison Plot for Subject 3 Setting 6.}
	\label{fig:wuren_3}
\end{figure*}

\begin{figure*}[h!]
	\centering
	\includegraphics[width=1\linewidth]{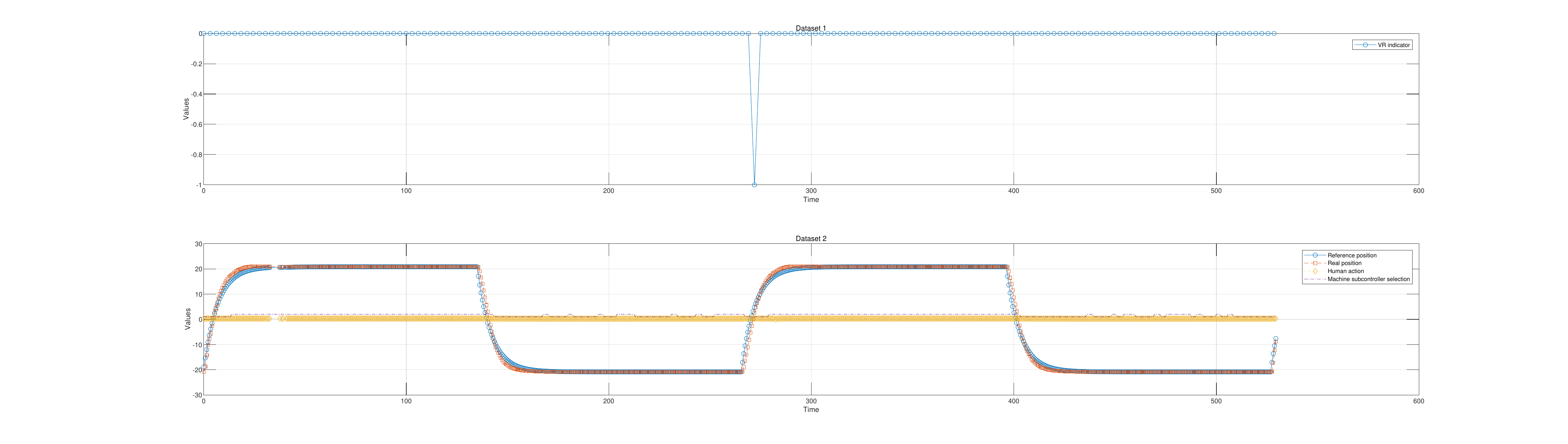}
	\caption{Comparison Plot for Subject 3 Setting 8.}
	\label{fig:wuren_4}
\end{figure*}

\begin{figure*}[h!]
	\centering
	\includegraphics[width=1\linewidth]{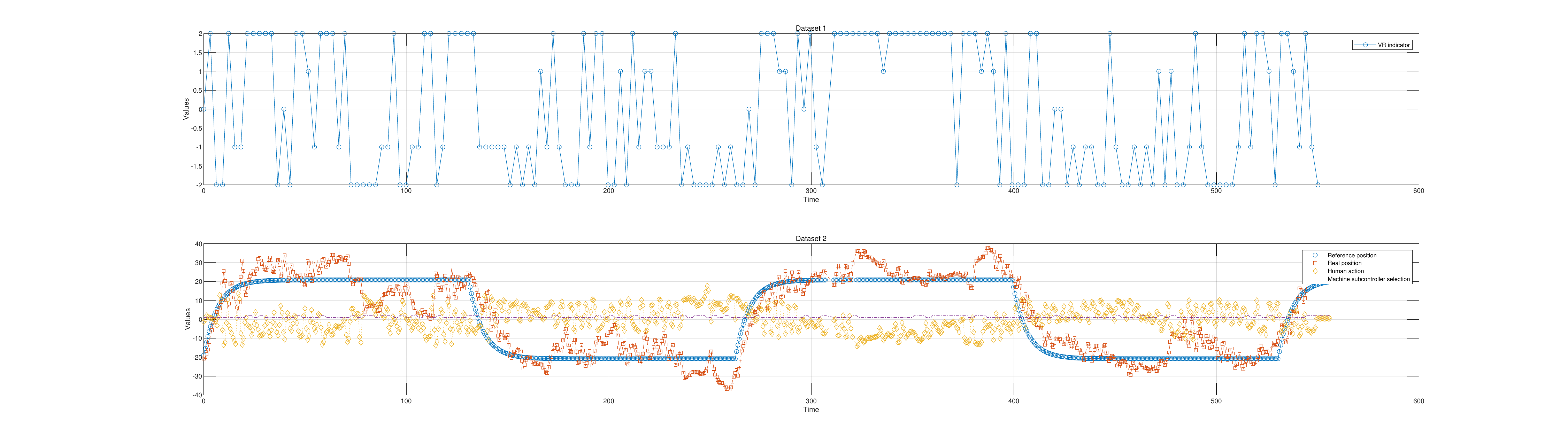}
	\caption{Comparison Plot for Subject 4 Setting 2.}
	\label{fig:liyaqi_2}
\end{figure*}

\begin{figure*}[h!]
	\centering
	\includegraphics[width=1\linewidth]{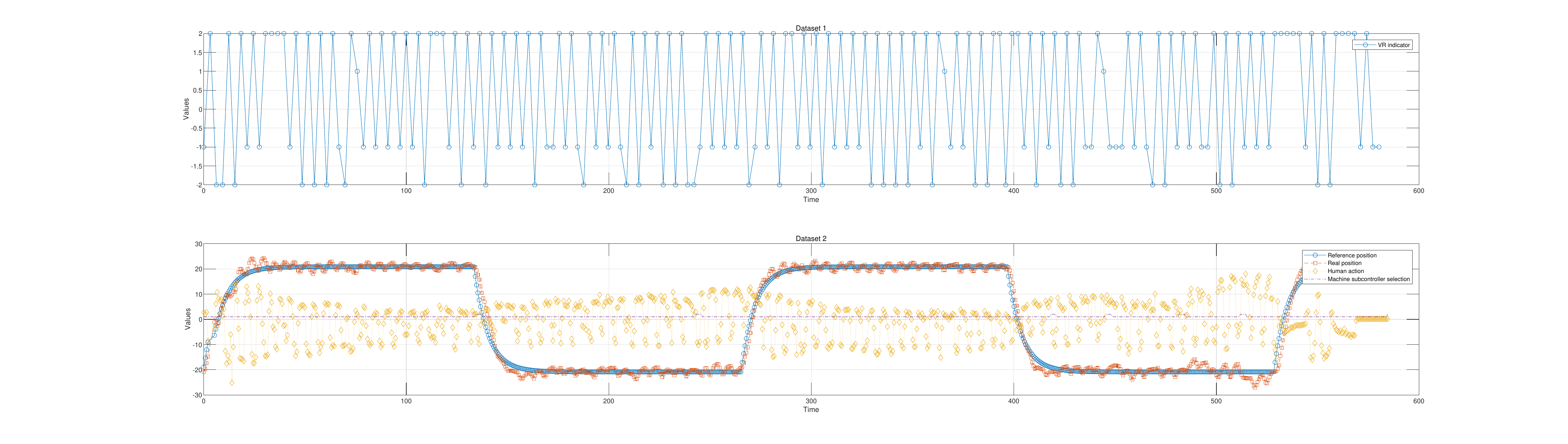}
	\caption{Comparison Plot for Subject 4 Setting 4.}
	\label{fig:liyaqi_4}
\end{figure*}

\begin{figure*}[h!]
	\centering
	\includegraphics[width=1\linewidth]{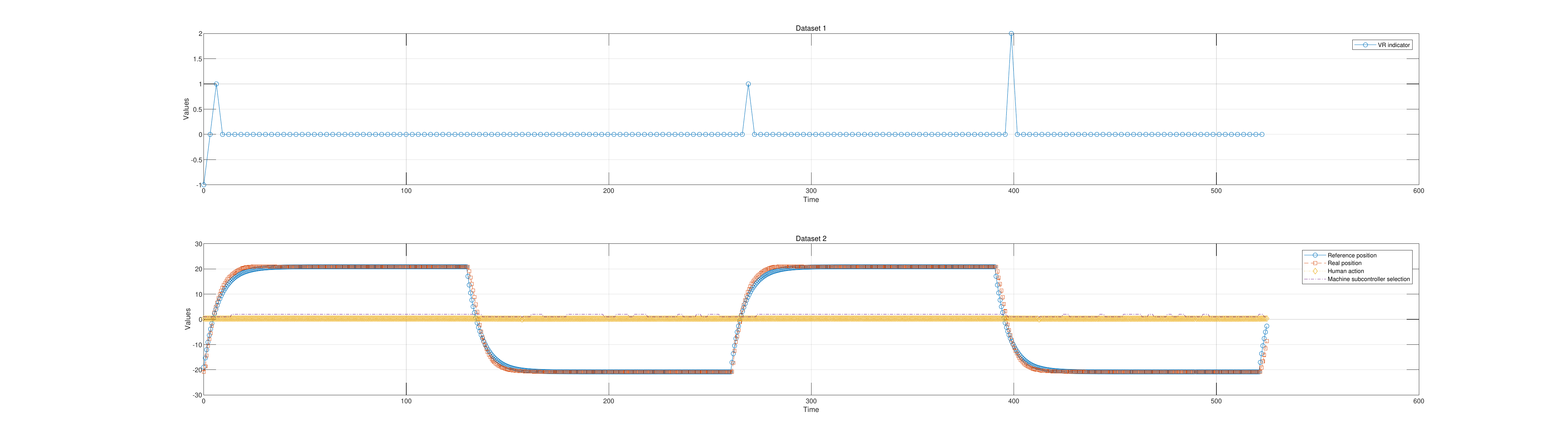}
	\caption{Comparison Plot for Subject 4 Setting 6.}
	\label{fig:wuren_5}
\end{figure*}

\begin{figure*}[h!]
	\centering
	\includegraphics[width=1\linewidth]{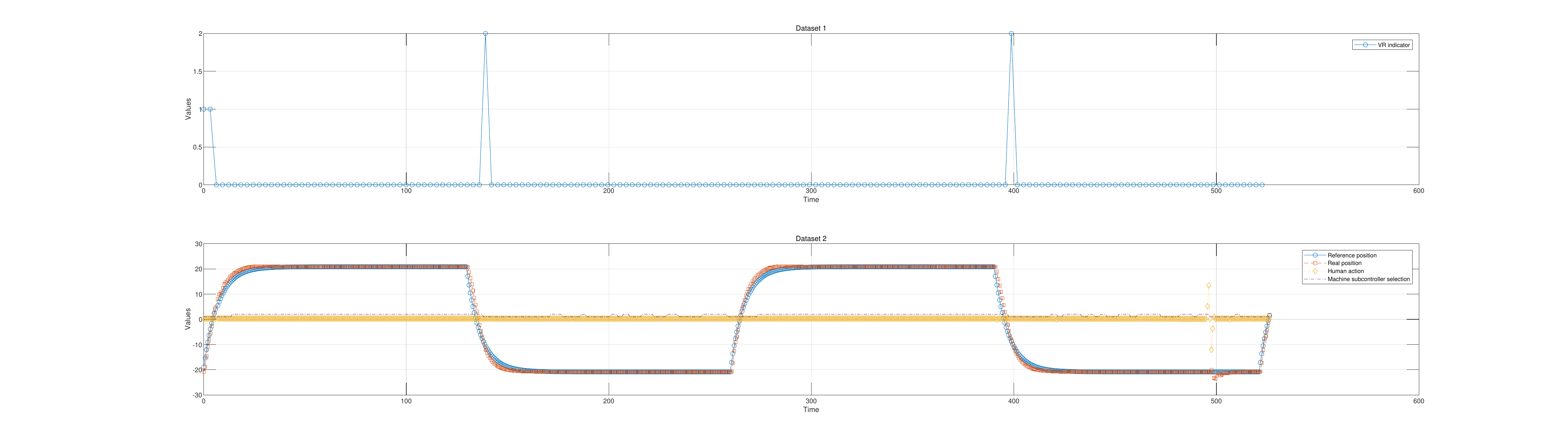}
	\caption{Comparison Plot for Subject 4 Setting 8.}
	\label{fig:wuren_6}
\end{figure*}

\begin{figure*}[h!]
	\centering
	\includegraphics[width=1\linewidth]{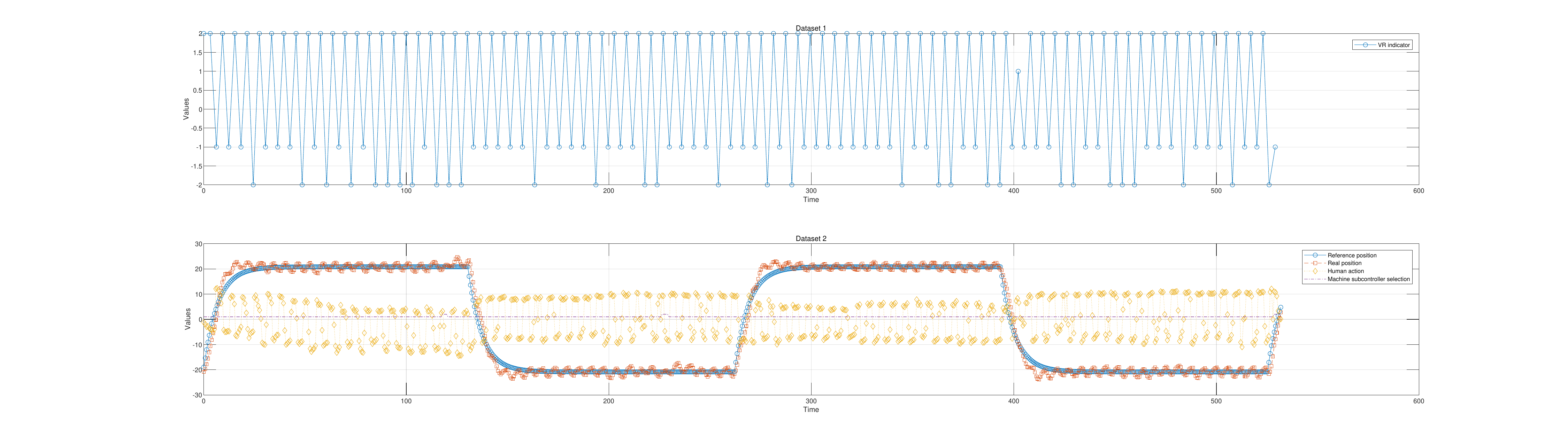}
	\caption{Comparison Plot for Subject 5 Setting 4.}
	\label{fig:suweidong_4}
\end{figure*}

\begin{figure*}[h!]
	\centering
	\includegraphics[width=1\linewidth]{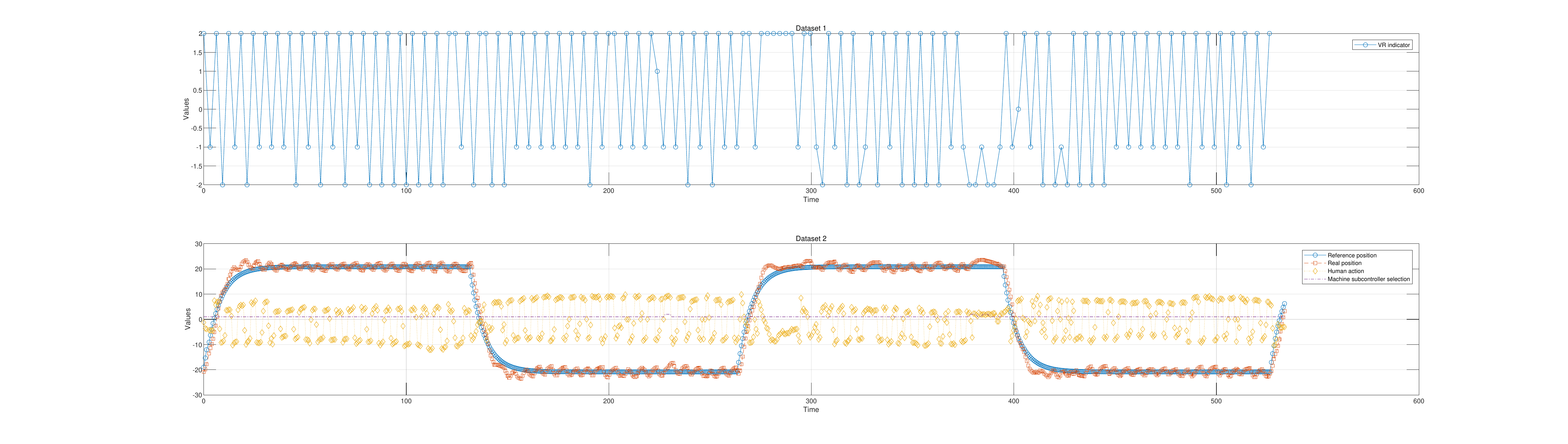}
	\caption{Comparison Plot for Subject 5 Setting 4-2.}
	\label{fig:suweidong_4_2}
\end{figure*}

\begin{figure*}[h!]
	\centering
	\includegraphics[width=1\linewidth]{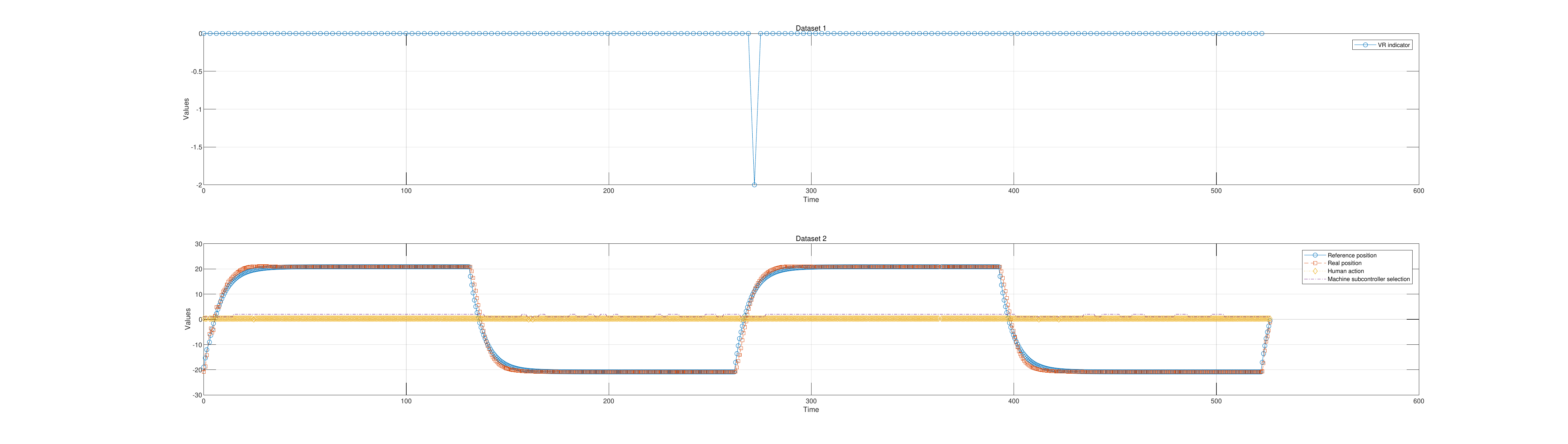}
	\caption{Comparison Plot for Subject 5 Setting 6.}
	\label{fig:wuren_7}
\end{figure*}

\begin{figure*}[h!]
	\centering
	\includegraphics[width=1\linewidth]{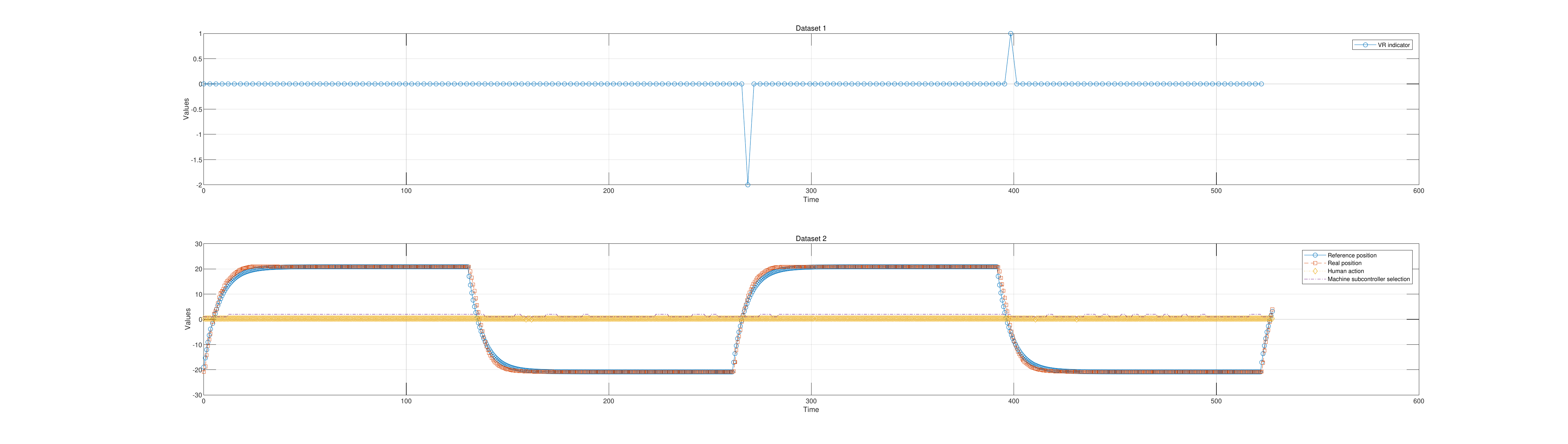}
	\caption{Comparison Plot for Subject 5 Setting 8.}
	\label{fig:wuren_8}
\end{figure*}

\begin{figure*}[h!]
	\centering
	\includegraphics[width=1\linewidth]{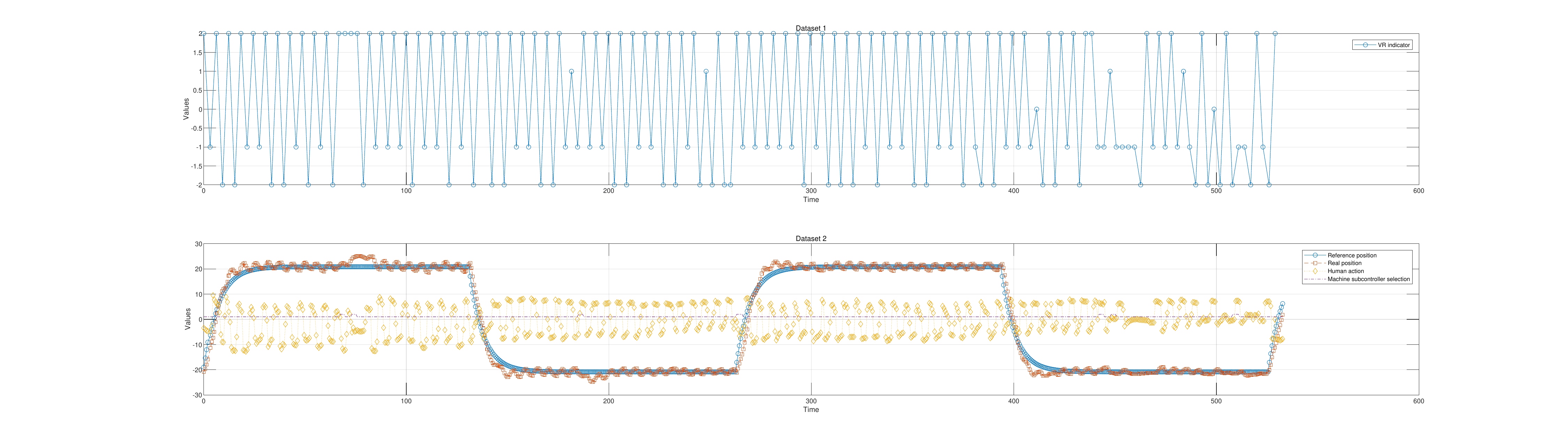}
	\caption{Comparison Plot for Subject 6 Setting 4.}
	\label{fig:tongyuhao_4}
\end{figure*}

\begin{figure*}[h!]
	\centering
	\includegraphics[width=1\linewidth]{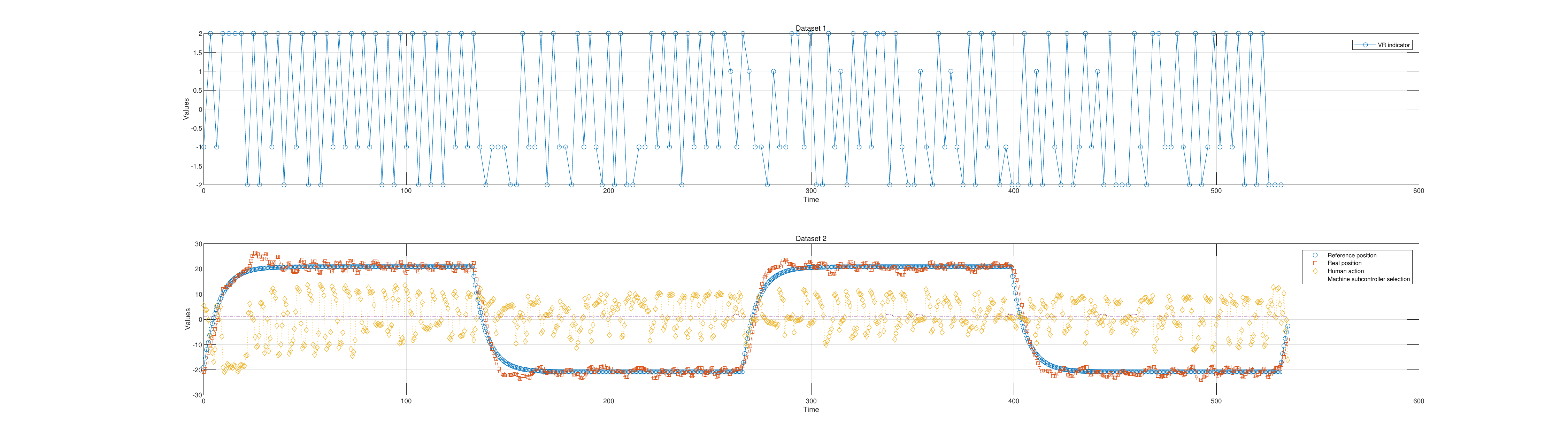}
	\caption{Comparison Plot for Subject 7 Setting 4.}
	\label{fig:wanghongwei_4}
\end{figure*}

\begin{figure*}[h!]
	\centering
	\includegraphics[width=1\linewidth]{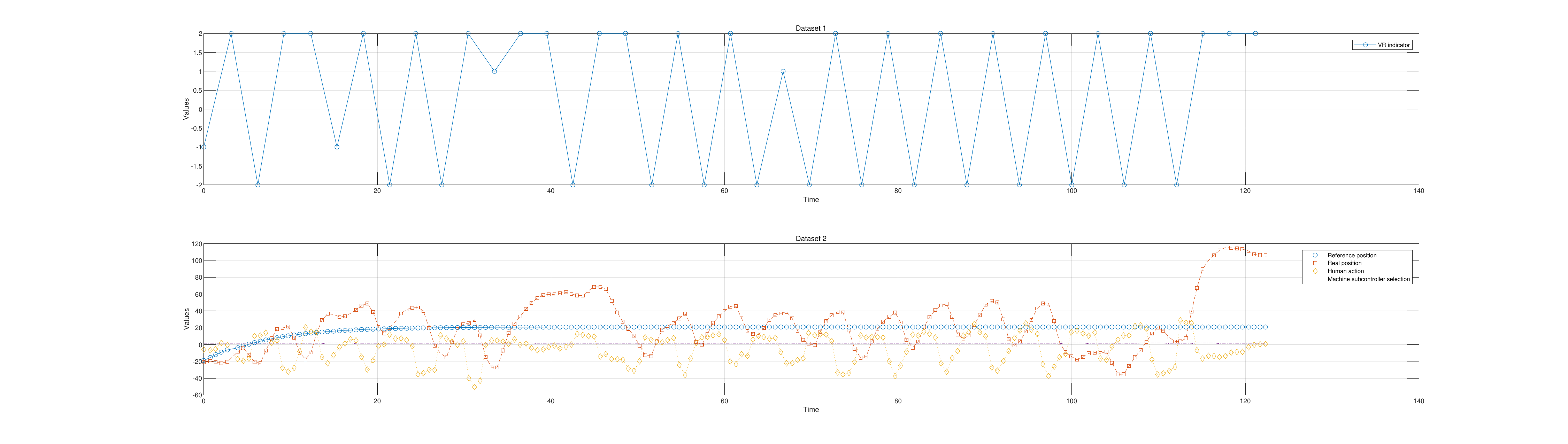}
	\caption{Comparison Plot for Subject 13 Setting 2.}
	\label{fig:yuruize_2}
\end{figure*}

\begin{figure*}[h!]
	\centering
	\includegraphics[width=1\linewidth]{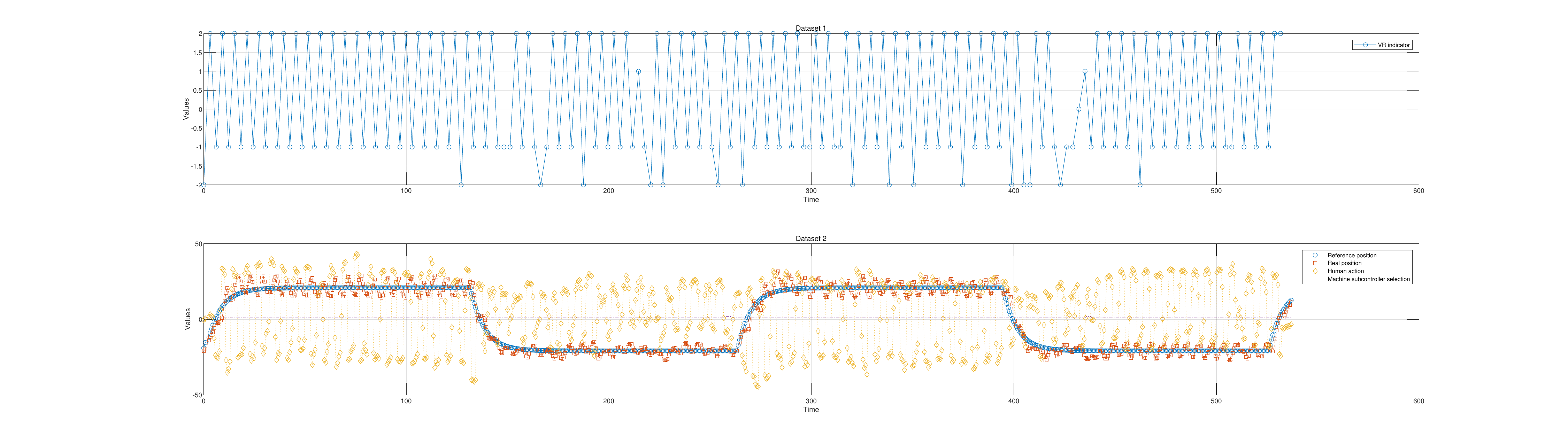}
	\caption{Comparison Plot for Subject 13 Setting 4.}
	\label{fig:yuruize_4}
\end{figure*}

The detailed experimental results are shown in Figures \ref{fig:anyang_2} to \ref{fig:yuruize_4}. To investigate the effectiveness of the reward function selection vs the testing outcomes, the testing results of Subject 1 are presented in Figures \ref{fig:anyang_2} to \ref{fig:anyang_8} as the patient input-output biofeedback model used in the simulation environment was based on Subject 1. These figures clearly indicate that the rehabilitation effects can be achieved based on the specifically designed reward functions, demonstrating the desired consistency with the simulated values (See Table \ref{table:pid_models_rewards}).

\begin{table}[ht]
	\centering
	\caption{Comparison of MSE Values of Subject 1 for Selected Settings}
	\begin{tabular}{|c|c|c|c|}
		\hline
		\textbf{Setting} & \textbf{Human Action (MSE)} & \textbf{Tracking Error (MSE)} & \textbf{Value} \\ \hline
		2 & 502.25 & 308.80 & 1940.50 \\ \hline
		4 & 468.17 & 4.57 & 2738.69 \\ \hline
		6 & 419.35 & 2.29 & 908.92 \\ \hline
		8 & 460.84 & 1.46 & 1626.40 \\ \hline
	\end{tabular}
	\label{tab:comparison_mse}
\end{table}

Specifically, in Figures \ref{fig:anyang_2} and \ref{fig:anyang_4} (corresponding to Settings 2 and 4 of Table \ref{table:pid_models_rewards} respectively), it is evident that Subject 1 actively participated in the exercise, as indicated by the higher level of human action (yellow marked) because the designed reward function is encouraging the patients' active participation. On the other hand, the human action in Figures \ref{fig:anyang_6} and \ref{fig:anyang_8} (corresponding to Settings 6 and 8 respectively) is lower because the designed reward function for these two settings is more emphasize the tracking accuracy rather than patient's participation. We can also see this results via MSE value: the human action MSE for setting 2 (Figure \ref{fig:anyang_2}) and setting 4 (Figure \ref{fig:anyang_4}) are 502.25 and 468.17 respectively. They are higher than those for Setting 6 (Figure \ref{fig:anyang_6}) and Setting 8 (Figure \ref{fig:anyang_8}) (See Table \ref{tab:comparison_mse}).

Furthermore, in Figures \ref{fig:anyang_2} and \ref{fig:anyang_4} (corresponding to Settings 2 and 4), it is evident that the system has much bigger tracking error in the exercise because the designed reward function is encouraging the patients' active participation rather than tracking accuracy. On the other hand, the tracking accuracy in Figures \ref{fig:anyang_6} and \ref{fig:anyang_8} (corresponding to Settings 6 and 8 respectively) is much lower because the designed reward function for these two settings is more emphasize the tracking accuracy rather than patient's participation. We can also see this results via MSE value: the MSE for tracking error of setting 2 (Figure \ref{fig:anyang_2}) and setting 4 (Figure \ref{fig:anyang_4})  are higher than those for Setting 6 (Figure \ref{fig:anyang_6}) and Setting 8 (Figure \ref{fig:anyang_8}) (See Table \ref{tab:comparison_mse}).

At the same time, we should also pointed out that under the same reward function, in training, Setting 4  achieved a value of $2738.69$, which is significantly higher than the value ($1940.50$) achieved by setting 2. Consistently, in experiments, the overall tracking error for setting 2 (Figure \ref{fig:anyang_2}) is much higher than that in Setting 4 (Figure \ref{fig:anyang_4}).

Similarly, under the same reward function (more emphasize on tracking accuracy), in simulation, setting 8 (Figure \ref{fig:anyang_8}) achieved a value of $1626.40$, compared to $908.92$ for setting 6 (Figure \ref{fig:anyang_6}). Consistently, in testing, the overall tracking error for setting 8 (Figure \ref{fig:anyang_8}) is lower than that for setting 6 (Figure \ref{fig:anyang_6}). 

Based on this analysis, we can conclude that for Subject 1, the training  values achieved for different settings is consistent with the training results in the simulation environment, as shown in Table \ref{table:pid_models_rewards}.

%
%

For other subjects, the overall consistency between simulation and testing is similar to that of Subject 1. Particularly for the reward function $r_{m} + 8r_{c} + r_{e}$ (encouraging robot assistance, resembling a passive rehabilitation mode), which corresponds to settings 5-8, the experimental results for different subjects are quite similar, as the effects of the patient are greatly reduced (see Figures \ref{fig:wuren_1}-\ref{fig:wuren_8} for details).

However, for the reward function $r_{m} + r_{c} + 5r_{e}$ (encouraging patients' active participation), which corresponds to settings 1-4, if the set of sub-controllers is not adjusted, different subjects may achieve varying training results. To observe intra-subject differences, consider Subject 13's experimental results for setting 2 (Figure \ref{fig:yuruize_2}) and setting 4 (Figure \ref{fig:yuruize_4}). Comparing these figures with Figures \ref{fig:anyang_2} and \ref{fig:anyang_4} for Subject 1, the differences are apparent. The actions from different patients can vary significantly (for setting 2, Subject 13 could not complete the exercise due to a much stronger biofeedback reaction compared to Subject 1).

\section{Conclusion}

In conclusion, this study addresses the challenges in robot-assisted ankle rehabilitation by introducing a Dual-Agent Multiple Model Reinforcement Learning (DAMMRL) framework, leveraging multiple model adaptive control (MMAC) and co-adaptive control strategies. Traditional single-model approaches have struggled to accurately model the complexities of human-machine interactions due to the dynamic and non-linear nature of human behaviour. By employing a multiple model strategy, we approximate complex human responses using a suite of simple sub-models tailored to varying levels of patient incapacity, thus overcoming the limitations posed by the curse of dimensionality.

The proposed framework's efficacy is demonstrated across real-world and simulated environments, showcasing its robustness and versatility. Rigorous evaluation with 13 healthy young subjects yielded promising results, affirming the anticipated benefits of this approach. This research not only introduces a novel paradigm for robot-assisted ankle rehabilitation but also sets the foundation for future studies focusing on adaptive, patient-centred therapeutic interventions, thereby enhancing the effectiveness and personalisation of rehabilitation processes through advanced reinforcement learning techniques.

\bibliographystyle{IEEEtran}
\bibliography{VO2}

\end{document}